\documentclass{article}

\usepackage[preprint]{neurips_2024}

\usepackage[utf8]{inputenc} % allow utf-8 input
\usepackage[T1]{fontenc}    % use 8-bit T1 fonts
\usepackage{hyperref}       % hyperlinks
\usepackage{url}            % simple URL typesetting
\usepackage{booktabs}       % professional-quality tables
\usepackage{amsmath}       % blackboard math symbols
\usepackage{amsfonts}       % blackboard math symbols
\usepackage{dsfont}
\usepackage{nicefrac}       % compact symbols for 1/2, etc.
\usepackage{microtype}      % microtypography
\usepackage{xcolor}         % colors
\usepackage{duckuments}
\usepackage{graphicx}
\usepackage{natbib}
\usepackage{listings}
\usepackage[most]{tcolorbox}

\newcommand{\systemname}[0]{\textsc{Formal Disco}}

\title{\systemname{}: Scalable Open-Ended Generation of Formally Verified Programs}

\author{%
  Gabriel Poesia$^{1}$ \quad
  Simon Henniger$^{2}$ \quad
  Tzu-Han Hsu$^{2}$ \quad
  Yilun Du$^{1,2}$ \quad
  Nada Amin$^{1,2}$ \\
  $^1$Kempner Institute, Harvard University \\
  $^2$School of Engineering and Applied Sciences, Harvard University \\
  Corresponding authors: 
  \texttt{gabriel\_poesia@fas.harvard.edu}, 
  \texttt{namin@seas.harvard.edu}
}

\lstdefinestyle{promptstyle}{
  basicstyle=\ttfamily\scriptsize,
  breaklines=true,
  breakatwhitespace=false,
  columns=fullflexible,
  keepspaces=true,
  showstringspaces=false,
  tabsize=2,
  frame=none,
}
\lstdefinestyle{codestyle}{
  basicstyle=\ttfamily\scriptsize,
  breaklines=true,
  breakatwhitespace=false,
  columns=fullflexible,
  keepspaces=true,
  showstringspaces=false,
  tabsize=2,
  frame=none,
}
\newtcolorbox{promptbox}[1]{
  colback=gray!5, colframe=gray!60!black, fonttitle=\bfseries,
  title=#1, breakable, boxrule=0.5pt, arc=1mm, left=2mm, right=2mm,
}
\newtcolorbox{codebox}[1]{
  colback=blue!3, colframe=blue!40!black, fonttitle=\bfseries,
  title=#1, breakable, boxrule=0.5pt, arc=1mm, left=2mm, right=2mm,
}

\newif\ifdraft
\drafttrue

\makeatletter
\def\@noticestring{Preprint.}
\makeatother

\begin{document}

\maketitle

\begin{abstract}
  The cost of producing code is rapidly diminishing with increasingly capable AI agents,
  while quality assurance of generated programs has not kept pace.
  Formal verification provides the strongest possible guarantees,
  but the ability of AI models to work with verification-aware languages is hindered by the scarcity of human-written examples of programs in those languages.
  To tackle this prevalent data scarcity issue, we propose \systemname: a distributed system for coordination of
  LLM-based workers that can be easily applied to open-ended synthetic data generation at scale.
  We use \systemname{} to share tasks and programs between three classes of workers: ``initiators'', which
  read random READMEs from open-source repositories and documentation snippets to sketch a related verified program,
  ``fixers'' which take compiler and verifier feedback and attempt to resolve issues, and ``extenders'' that take
  working programs and propose patches to expand them.
  \systemname{} records all agent-generated traces and uses them both for initial distillation from a stronger model as well as self-improvement.
  We also propose a principle of maximum entropy for synthetic program generation, and use entropy maximization via iterative supervised fine-tuning to learn to generate increasingly diverse programs over time.
  We release large datasets of synthetic verified programs in three languages --- Dafny, Verus, and Frama-C ---,
  and fine-tune open models for verification-relevant tasks, often matching or exceeding the performance of several proprietary models.
  Overall, our work offers a path to create synthetic data at scale
  for formal reasoning domains and overcome the long-standing data barrier.

\end{abstract}

\section{Introduction}

AI coding agents are rapidly changing the practice of software engineering \cite{robbes2026agentic}.
Frontier AI agents at scale have been shown to author complete, highly complex
code bases spanning tens to hundreds of thousands of lines of code, such as the C compiler implemented by Anthropic's Claude Code \cite{anthropic2026buildingccompiler}.
While the cost of \emph{writing code} is rapidly diminishing, we still lack means to verify the correctness of
AI-written programs at scale: testing can only cover finitely many cases, and manual code review is infeasible at this new pace.
But bugs can be extremely costly --- a bug in a compiler, for instance, can carry over to all
programs compiled by the compiler, with potential catastrophic consequences downstream. How can we ensure that programs behave as intended, so we can rely on AI-written code?

Formal program verification provides the strongest possible form of correctness assurance \cite{mugnier2025impact,leino2010dafny,lattuada2023verus,lattuada2024verus}.
A formally verified program comes with both a logical specification of its behavior as well as a proof that a given implementation
satisfies the specification. If AI agents write verified programs, humans can check correctness by writing or inspecting a formal specification,
but can then trust a symbolic verifier to judge whether the implementation satisfies the given specification.

Unfortunately, current LLMs, especially open-weight models, are still limited in their ability to write verified programs,
or even assist humans with verification-related tasks (such as annotating loops with invariants). This is to be expected given the lack
of human-written examples of programs in most languages supporting formal verification. For instance, while there are millions of Python
repositories on GitHub, the largest existing public dataset of programs in Dafny --- a verification-aware programming language ---
is DafnyBench \cite{loughridge2024dafnybench}, containing less than 800 Dafny files collected and deduplicated from public GitHub repositories.
% python repository: xxxM

In this paper, we tackle the data scarcity barrier that has hindered progress in AI for verified programming using LLMs themselves
to generate synthetic data at scale, in an \emph{open-ended fashion}. By open-ended, we imply that our primary goal is only to generate a large volume of new, \emph{diverse} verified programs in formal languages, including specifications, implementations, and proofs,
without being directed a priori towards a particular downstream task or application.
Ultimately, we then use these synthetic programs in a task-directed manner to train open models for verification-relevant tasks, evaluating
them using existing human-written programs.

We operationalize this idea introducing \systemname{}, a distributed system for LLM agents to interact via a shared ``agenda'',
which maintains tasks and a database of objects (e.g., programs). To generate synthetic programs at scale, we implement three
classes of agents: ``initiators'', that sample READMEs from random GitHub repositories and snippets from the language's documentation and sketch an initial program in a given language
that is related to the theme of that README, ``fixers'' that take programs that fail compilation or verification and attempt to patch them
to fix errors, and ``extenders'' that generate patches to expand already working programs, making them grow in complexity.
Random READMEs provide a useful source of theme entropy, which we observe to collapse if the ideas for verified programs come entirely
from LLMs themselves with no external signal. In turn, fixers and extenders allow us to get programs that are significantly more complex than
initial programs, allowing the initiator to avoid being overly ambitious while still ultimately producing complex programs.

\systemname{} records traces from all agents in the agenda, allowing us to both bootstrap open models via distillation from stronger proprietary
models, as well as to self-improve the agents at the relevant generation tasks, using the verifier as a reward signal. Moreover, we introduce a framework of \emph{entropy maximization} that allows us to not only improve agents at generating programs that compile and pass verification, but that are also \emph{increasingly diverse} in entropy metrics derived from logical specifications, annotations, loop structures and proofs for lemmas.

We use \systemname{} to release large public datasets of programs in three languages: Dafny, Frama-C, and Verus --- a recent verification framework for the
Rust programming language. Our datasets, dafny-disco, framac-disco and verus-disco, contain over 100k complete verified programs --- for Dafny and Frama-C, they have orders of magnitude more programs than existing public datasets. Our conceptual framework based on feature entropy allows us to compare \emph{datasets} of programs with each other, shedding light on their quality beyond just their raw size, which is a poor metric to assess datasets in general \cite{gupta2021data}.
We train and evaluate open models on our datasets for verification-related tasks in all languages, including annotating methods with assertions
and invariants so that verifiers can prove the specification holds, and proving lemmas. Overall, training on \systemname{} data yields substantial improvements to Qwen 2.5-Coder 32B, in some cases making it competitive with Claude 4.5 Opus.

In summary, we make the following contributions:

\begin{itemize}
  \item We introduce \systemname{}, a language-agnostic framework for coordinating LLM-based workers around tasks and an object database,
  collecting traces for distillation and self-improvement. We instantiate it with agents for open-ended generation of verified programs.

  \item We propose a conceptual framework of \emph{entropy maximization} over program features for synthetic program datasets, allowing us to compare datasets and improve towards both task success rates and increased diversity of the generated datasets.

  \item We release large datasets of programs in three verification-aware languages --- Dafny, Verus and Frama-C ---, generated synthetically with \systemname{}, and analyze the resulting datasets in relation to existing ones using feature entropy framework.

  \item We train open models using our data for verification-relevant tasks in Dafny and Verus, and find that training on our synthetic datasets improves model performance on verifying human-written programs, in some cases reaching performance competitive with Claude 4.5 Opus.
\end{itemize}

\section{Related Work}

\textbf{Formally verified code generation with LLMs.} A growing line of work aims to address the reliability concerns of LLM program synthesis through formal verification \cite{loughridge2024dafnybench,yang2025autoverus}: having models produce not just code but machine-checkable specifications and proofs, whether in interactive theorem provers (e.g., Lean, Isabelle) or in auto-active verification languages such as Dafny~\cite{leino2010dafny} and Verus~\cite{lattuada2023verus}, which target executable programs directly~\cite{misu2024, yang2025autoverus, sun2024clover}.

\textbf{LLMs and data-scarce programming languages.} A central obstacle for this endeavour is data scarcity. Verification-aware languages are extreme cases of ``low-resource'' languages, with relatively little representation in public datasets compared to popular languages like Python, and LLM performance is known to degrade sharply on languages underrepresented in training data~\cite{cassano2023multiple}, and while knowledge-transfer techniques that translate training data from high-resource languages have proven effective for conventional low-resource languages~\cite{cassano2024multiplt}, verified code adds the further requirements of formal specifications and proofs, for which much less natural data exists.

\textbf{Synthetic verified program generation.} Several recent approaches, mainly focused on Dafny and Verus, have used synthetic data synthesis for tackling the data scarcity issue in AI for formal verification. \textbf{Verus.} Chen et al.~\cite{chen2026automatedproofgenerationrust} describe the problem as multi-level data scarcity: there is a shortage not only of proofs, but also of \emph{formal specifications} and even of Rust programs written in Verus-supported syntax. Their dataset, SAFE, was constructed used OpenAI's GPT-4o to translate tens of thousands of small Python and Rust programs from popular code-synthesis datasets into Verus-compatible Rust, and then generate specifications and proofs through inference and iterative fine-tuning of open models. This approach, however, is limited by its original seed programs. The follow-up work on VeruSyn \cite{di2026reducingcostsproofsynthesis} started with SAFE's seed dataset and scale up synthesis up to a dataset of 6.9M programs. VeruSyn expands the seed data distribution using ``tutorial-based synthesis'' to explicitly broaden coverage of Verus-specific features beyond what the seed data contains (SAFE, for instance, contained no instances of lemmas --- known as ``proof functions'' in Rust --- despite this being a commonly necessary feature for proving complex specifications). \textbf{Dafny}. There has also been parallel approaches for synthetic training data generation in Dafny, including ATLAS \cite{baksys2026atlasautomatedtoolkitlargescale}, which focused on using seed programs in a higher-resource language, such as Python, and attempting to reimplement them into Dafny (which includes adding suitable specifications and proofs, not present in the original program), as well as the DafnySynth component in dafny-annotator \cite{poesia2024dafnyannotatoraiassistedverificationdafny}, a fully synthetic dataset generated by scheduling GPT-4o to sketch, implement and extend new Dafny programs.
Compared to these, \systemname{} is the first synthetic program generation method that is shown to be language-agnostic --- we apply it to Dafny, Verus and Frama-C, and adding new languages is simple. Moreover, \systemname{} can directly optimize and improve the diversity of generated programs across iterations, overcoming a common worry that self-training LLMs on their own outputs can collapse diversity \cite{herel2024collapse,shumailov2024ai}.

\textbf{Synthetic data in formal mathematics.} Beyond verified programming, synthetic data has driven recent progress in formal theorem proving, where proof checkers provide the same scalable correctness signal that program verifiers give us. AlphaGeometry~\cite{trinh2024solving} reached near gold-medalist performance on olympiad geometry after pre-training on 100M synthetically generated theorems and proofs, while DeepSeek-Prover~\cite{xin2024deepseek} and Goedel-Prover~\cite{lin2025goedel} grow large corpora of Lean proofs through cycles of autoformalization, proof search, verifier filtering and re-training --- a loop analogous to our distillation and self-improvement pipeline. These efforts, however, generate proofs for externally given (or translated) statements; \systemname{} instead generates specifications, implementations and proofs jointly and open-endedly, as has been explored in formal mathematics \cite{poesia2024learning,dong2025stp}, here with dataset diversity as an explicit objective.

\textbf{Open-endedness and diversity in synthetic data generation.} \systemname{} descends from classical open-ended discovery systems like AM and Eurisko~\cite{lenat1984}, a lineage recently revived by LLM-based open-ended agents and generators~\cite{wang2023voyager}. In synthetic data generation more broadly, a recurring lesson is that generation must be \emph{seeded} with external sources of randomness to avoid collapsing onto few themes: for instance, TinyStories conditions story generation on random word combinations~\cite{eldan2023tinystories} and phi injects randomness into prompts for generating synthetic textbooks~\cite{li2023textbooks}. Our READMEs and documentation snippets play this same role for verified programs. But where prior work treats diversity as a qualitative design goal, our entropy-maximization framework turns it into a metric that can be tracked and directly optimized.

\section{\systemname}

\begin{figure}
  \centering
  \includegraphics[width=\textwidth]{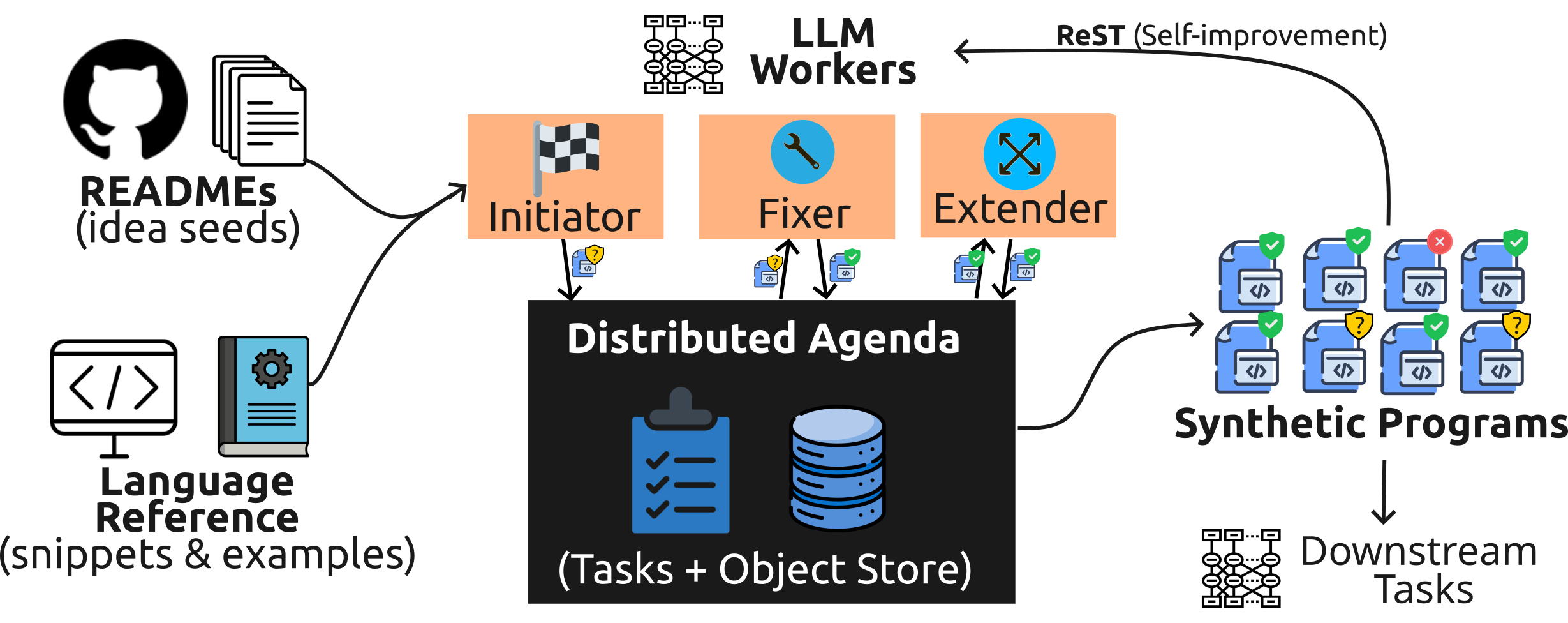}

  \caption{Overview of \systemname. The system is centered around a distributed agenda, where LLM-based workers collaborate on completing tasks and store objects --- mainly synthetic programs being written, extended and repaired. The \emph{Initiator} worker sketches initial programs in the given verification-aware language, including specifications and proofs, taking ideas from README files sampled from open-source repositories and snippets and examples from the programming language's reference. Programs might fail compilation or verification; a \emph{Fixer} worker then writes patches to attempt to fix errors. Working programs grow in complexity by patches written by the \emph{Extender} worker. Resulting synthetic programs can then be used both to self-improve the underlying LLM, as well as to generate training data for downstream tasks.}
  \label{fig:overview}
\end{figure}

In this paper, our main goal is to enable scalable synthetic generation of programs in languages where a formal verifier is available but few human-written programs exist --- these both include languages like Dafny, which have existed for many years but have not reached the mainstream, as well as more recent languages and frameworks, like Verus, an ongoing effort to enable formal verification of Rust programs. 
%Ultimately, \systemname{} also enables us to picture how languages of the future, which by design will be released with few existing human-written programs in them, might be released together with large-scale (synthetic) datasets so that even early users can benefit from AI assistance, which programmers are learning to increasingly rely on and which might become a strong factor to drive (or hinder) future adoption of new tools.
% SH: I like this, why commented out?
But what makes \emph{a high-quality dataset of programs?} A useful quality metric for a \emph{dataset} can both serve to guide the design of the overall system as well as an objective function for direct optimization. Thus, the first goal of this section is to determine such a metric. We here present a framework analogous to the principle of maximum entropy, a classical principle in statistics for selecting models, but instead applied to guide \emph{data generation}. Then, we describe the agents we set up in \systemname{} for open-ended generation of verified programs, and finally operationalize self-improvement of the agents both for task success rate and entropy maximization over program features, learning to generate correct and increasingly diverse programs over time.

\subsection{A principle of maximum entropy for synthetic program generation}

What makes a good dataset of programs? Several intuitions have been used by prior work on synthetic program generation in Verus --- notably in the SAFE and VeruSyn synthetic datasets --- as a guide. For one, larger datasets tend to be more useful for training LLMs. But size alone is a poor metric: for instance, it would be possible to create an unbounded number of trivial variations of a given seed program to generate arbitrarily many more, and yet we would fail to generate useful training signal for most purposes. Another intuitive property is coverage over language features. For instance, programs in the SAFE dataset do not have any lemmas, generally necessary to prove deeper properties that the SMT solver alone cannot discharge. Several other Verus features are also entirely absent from SAFE, an issue that VeruSyn mitigated by using samples from the Verus documentation to generate seed programs. Other desirable properties are related to the complexity of the programs: ideally we would want programs with various loop structures, requiring many logical operations to prove specifications, functions with complex implementations, and so on. \emph{Is there any underlying principle behind all of these desiderata?}

We propose that all of these desirable properties --- including dataset size, coverage over language features, and various notions of program complexity --- can be achieved by a unified goal of seeking \emph{entropy over program features}. More precisely, let $\mathcal{P}$ be the set of valid programs in a given language. A (discrete) program feature is then a function $f : \mathcal{P} \rightarrow \mathcal{V}_f$, where $\mathcal{V}_f$ is an enumerable set of possible values that $f(p)$ can take. Given a finite dataset $\mathcal{D} \in 2^{\mathcal{P}}$, feature $f$ gives rise to a probability distribution $P^{\mathcal{D}}_f(v)$ over the value of $f(p)$ for a program $p$ drawn uniformly from $\mathcal{D}$, where

\begin{equation}
P^{\mathcal{D}}_f(v) = \frac{1}{\mathcal{D}} \sum_{p \in \mathcal{D}} \mathds{1}[f(p) = v]    
\end{equation}

We define the entropy $H(f, \mathcal{D})$ of feature $f$ in a given dataset $\mathcal{D}$ to be the Shannon entropy of $P^{\mathcal{D}}_f$:

\begin{equation}
\label{eq:h}
    H(f, \mathcal{D}) = - \sum_{v \in \mathcal{V}} P^{\mathcal{D}}_f(v) \log P^{\mathcal{D}}_f(v)
\end{equation}

Intuitively, $H(f, \mathcal{D})$ serves as a measure of the diversity of values of $f$ observed in $\mathcal{D}$. The key principle we build upon is that a \emph{good synthetic dataset of programs will maximize its entropy over a set of relevant program features}. This hypothesis unifies several intuitions on dataset quality:

\begin{description}
    \item[Larger datasets can have larger feature entropy.] The fact that the uniform distribution has higher entropy than any other probability distribution supported on a discrete set leads to a simple upper bound on feature entropy: $H(f, \mathcal{D}) \leq \log |\mathcal{\mathcal{D}}|$. This explains why larger datasets are typically better: they have a larger \emph{potential} for feature entropy.
    \item[Dataset size alone does not guarantee utility.] However, the raw size of a dataset does not guarantee it indeed has large entropy. For instance, a large dataset comprised of $10^{12}$ copies of the same program will have $H(f, \mathcal{D}) = 0$ for any $f$.
    \item[High entropy in complexity metrics requires complex programs.] One desirable property of $\mathcal{D}$ is that it should contain complex programs. This is also captured by having high $H(f_c, \mathcal{D})$ over some $f_c$ that is a program feature related to program complexity (e.g., the maximum loop nesting depth in the entire program). If all programs are simple or trivial (e.g., if there are no loops), $f_c$ will collapse into a small number of possible values (e.g., 0 maximum nesting depth if there are no loops in the corpus, or at most 1 if there are no nested loops). Thus, the absence of complex programs as measured by $f_c$ will be reflected in $\mathcal{D}$ having low $H(f_c, \mathcal{D})$.
    \item[Coverage over programming language features can be measured by entropy.] The desideratum that a rich dataset should cover the usage of a variety of relevant programming language features (e.g., all of the features appearing in its reference manual~\cite{di2026reducingcostsproofsynthesis}) can also be seen through the lens of entropy maximization. Here, the relevant program features $f_e(p)$ are the binary features checking whether program $p$ uses a given feature $e$ from the target programming language (for instance, whether $p$ uses lemmas, has \texttt{\#[trigger]} annotations, or ghost code in Rust). If no program in a dataset uses $e$, the entropy over this feature will be zero: maximizing this entropy will require introducing programs that use $e$ (and others that do not).
\end{description}

Although not framed using our framework, this principle of maximum entropy for synthetic datasets helps unify the motivation behind prior work on synthetic program generation. For instance, the recent VeruSyn dataset, of synthetic Rust programs, is much larger (thus has higher potential entropy) than its direct predecessor SAFE. Moreover, the authors in VeruSyn specifically noted that SAFE does not include any programs that use many relevant Verus features, such as lemmas, that are often found in real-world Verus code. This can be rephrased as the observation that SAFE has entropy 0 over relevant features indicating usage of various Verus features (e.g., $f_\texttt{lemma}(p) = \mathds{1}[p \text{ has lemmas}]$). Indeed, the qualitative contributions proposed in VeruSyn to improve upon its seed dataset SAFE are rendered \emph{measurable} via the lens of feature entropy and casting entropy maximization as the objective for dataset synthesis.

%For such languages,
%typically open models perform poorly at generating valid programs, given the scarcity of verified programs in public datasets. But every attempt to write
%or modify code can receive feedback from the verifier on whether it is able to prove the specifications stated in the program hold.
%Thus, we would like to use this feedback to both \emph{distill} successful examples from stronger models,
%as well as to \emph{self-improve} models used to implement workers as they find more successful examples.
%To these ends, as we describe below, all workers save distillation examples (prompts, LLM responses and verifier feedback)
%after all task attempts. These examples are stored as objects in the agenda under a ``distill'' virtual directory, and are
%used between subsequent runs of the whole system to train or improve open models at the worker tasks via supervised fine-tuning.

This objective will serve to inform the design of \systemname{} , a multi-agent system we introduce next, to generate synthetic verified programs: qualitatively, its agents and components (Section~\ref{sec:workers}) will be chosen to introduce entropy over relevant program features; quantitatively, dataset entropy will serve directly as an objective for self-improvement in Section~\ref{sec:self-improvement}.

\subsection{Open-ended verified program generation with LLM-based workers}
\label{sec:workers}

We now describe \systemname{}, a multi-agent system for synthetic program generation.
\systemname{}'s architecture is inspired by classical \emph{discovery systems} such as AM and Eurisko~\cite{lenat1984}, which aimed at automating mathematical discovery. Here, these ``discoveries'' will consist of formally verified programs, although the base system can in principle be adapted to other settings including mathematics. The whole system is illustrated in Figure~\ref{fig:overview}.
As those classical discovery systems, \systemname{} is centered around an ``agenda'',
which keeps track of both tasks and objects. Objects are identified by a path, mimicking a virtual file system --- for instance, programs are saved as objects (e.g., with path \texttt{programs/prog123.rs}), with their verification status and feedback attached as metadata. Tasks
can point to object dependencies and have a type and other data about the work they require. This design is analogous in many ways to early discovery systems, except that the agents creating and editing objects, while working on tasks, are language model-based workers, rather than traditional heuristic programs. Furthermore, \systemname{}'s agenda is distributed: the agenda exposes a remote API
so various \emph{workers} can asynchronously create or claim tasks, as well as create and update objects.
This facilitates horizontal scaling, as new agents can enter the system at any time. In worker processes, we use a simple round-robin \emph{scheduler} that runs through a list of workers and invokes them in sequence for one unit of work each. Specifically, we implement the following three workers:

\begin{description}
    \item[Initiator:] Samples a random \emph{seed context} (described below) and prompts an LLM with the context to write a small, initial verified program in the target language, along with comments with ideas on how it can be extended. The generated program is saved to a new unique path as an agenda object. The initiator then invokes the compiler and verifier to collect feedback on the generated program. If the program compiles and verifies with no errors, it creates an ``extend'' task for this program and adds it to the agenda. If the program has any errors (e.g., in syntax, compilation or verification), the initiator creates a ``repair'' task in the agenda. One unit of work for the initiator generates exactly one new task in the agenda (and consumes none).
    \item[Fixer:] Finds and claims a task of type ``repair'' in the agenda, skipping its turn if none exists. It reads the program from the task dependency, as well as compiler/verifier errors saved in its metadata. It then prompts an LLM to first output a chain-of-thought reasoning about the errors and end with a diff such that should fix (at least some of) the reported errors when applied to the program. If the diff is in the correct format and can be applied, the program is updated in the agenda. The compiler/verifier is then invoked again: the task is marked completed if the program has no remaining errors, and a new task with type ``extend'' is created for this program. Otherwise, the repair task is left in the agenda, marked as ``attempted'' with an increased number of registered attempts. If a maximum number of attempts has already been reached in this task (3 in our runs), we instead mark the task as ``failed'' and remove it from the agenda.
    \item[Extender:] Finds and claims a task of type ``extend''. It reads the associated program from the agenda (which is known to compile and verify), and prompts an LLM to reason about a possible extension to the program (e.g., adding a new lemma, a new method, implementing something new building on what the program implements, etc). The LLM should also output a diff, that is then applied to the program, which is updated. If the new program passes compilation and verification, the extender creates a further ``extend'' task for this program. Otherwise, it creates a ``repair'' task for this program.
\end{description}

One \emph{run} in \systemname{} consists of starting an agenda and scheduling workers to complete a maximum of $K$ task attempts (up to $30,000$ in our experiments). Open-ended program generation thus starts with the Initiator, which receives a randomly sampled seed context. This context consists of two parts. First, we give the Initiator a sampled GitHub README.md file from a dataset we collected of 100k repositories from GitHub\footnote{These READMEs were sampled from Google's BigQuery GitHub database, and are generally (simply given the overall distribution of programs on GitHub) not coming from verification-related projects.}. As we show in the experiments, random READMEs allow us to significantly improve entropy of the generated corpus with respect to program ``themes'' (e.g., as indicated by the identifiers found in the programs). The second part of the context is a set of up to 2 randomly sampled snippets generated from the language's reference. We prepare these snippets beforehand to contain a description of a feature of the programming language paired with one usage example. This part of the seed helps increase entropy with respect to \emph{language feature coverage}: many features of Dafny or Verus, for instance, do not appear in the generated corpus if seed snippets from the reference are not present.

Note that the generation is open-ended in the sense that we allow the LLM workers to choose what programs will be about, with respect to both specification and implementation: we explicitly describe the READMEs and documentation snippets as seeds to potentially draw from, but do not set it as a requirement that the generated program has to attempt to be a faithful translation of what is described in the README (which is generally too ambitious of a requirement, given that most repositories hardly lead to clear related verified programs), or use the explained features at all costs. Still, the idea here is that simply aiming to obtain a diverse corpus of programs will lead to useful training data, especially for languages that lack it.

\subsection{Distillation and self-improvement via entropy maximization}
\label{sec:self-improvement}

So far, the goal of maximizing entropy over program features has been used only qualitatively as a guide for the design of \systemname. We now go further and use it to derive a specific objective for distillation and self-improving the three LLM-based workers described above.

In a \systemname{} run, every LLM call is also recorded to the agenda as a \emph{distillation example}, with its full prompt, LLM output, and the \emph{outcome} of the attempt to fulfill the worker's task. Specifically, all workers are deemed to succeed if the resulting program passes both compilation and verification with no errors. We find that existing open-weights LLMs, such as from the Qwen Coder family of models, are generally not yet capable in the verification-aware languages we use. Thus, for each language, we first perform initial runs using closed frontier models (e.g., Claude Opus and Sonnet) to collect initial \emph{distillation} data. Success rate even for these models is typically below $15\%$ for most worker tasks, which makes it expensive to exclusively rely on these models to produce a large-scale dataset. However, even a run with $10k$ attempts gives enough distillation data to allow open models to become useful. After this initial seed run, we run subsequent runs using open models. Between runs, we use the data collected so far for supervised fine-tuning (SFT), using LoRA.

What training examples should we use for SFT? Using \emph{all} collected examples, including failed ones, gives more data but cannot surpass the relatively low success rates that the seed model achieves. Thus, the choice of which examples to train on gives us a lever for allowing the LLM to iteratively improve at the worker tasks.

But what should ``improving'' mean here? If the only relevant signal is the compiler and verifier feedback, all workers can converge to succeed by generating only simple programs. To circumvent this issue, we use the framework of \emph{entropy maximization} to formulate an objective that to steer workers towards generating increasingly diverse programs, and use a simple ranking and filtering scheme (as in ReST~\cite{gulcehre2023reinforced} and RAFT~\cite{dong2023raft}) to optimize for this objective.

Specifically, let $f_1, \cdots, f_k$ be a set of program features over which we want the dataset we generate to have high entropy. For each $f_i$, the definition of $H(f_i, \mathcal{D})$ in Equation~\ref{eq:h} can be reformulated in terms of the contributions of each individual program in $\mathcal{D}$. Specifically, we define the \emph{$f$-surprisal} $S(f, p, \mathcal{D})$ of program $p$ relative to $\mathcal{D}$ as:

\begin{equation}
    S(f, p, \mathcal{D}) = -\log P^\mathcal{D}_{f}(f(p))
\end{equation}

Intuitively, the surprisal measures the ``rarity'' of $f_i(p)$ among programs in $\mathcal{D}$. With $S$, we can thus rewrite $H(f, \mathcal{D})$ as:

\begin{equation}
    H(f, \mathcal{D}) = \frac{1}{|\mathcal{D}|} \sum_{p \in \mathcal{D}} S(f, p, \mathcal{D})
\end{equation}

This decomposition of $H(f, \mathcal{D})$ in terms of the contributions of each program allows us to rank programs based on their contributions to the overall entropy. Let $\text{rank}(p, f, \mathcal{D}) \in [1, |\mathcal{D}|]$ be the order of $p$ in $\mathcal{D}$ when programs are sorted by their $f$-surprisal (with 1 being the program with the highest surprisal value; ties broken arbitrarily). Then, we define the \emph{minimum surprisal rank} ($\textit{msr}$) of $p$ as:

\begin{equation}
    \textit{msr}(p, \mathcal{D}) = \min_{f \in \{f_1, \cdots, f_k\}} \text{rank}(p, f, \mathcal{D})
\end{equation}

Then, in each iteration of SFT on the accumulated dataset, we only train on the top $1/3$ successful examples as ranked by the $\textit{msr}$ of the programs they generated, maintaining the fraction of examples of each prompt type available (``initiate'', ``repair'' and ``extend''). As we will show in our experiments, training only on the examples that contribute the most to the entropy of the generated dataset makes each batch generate corpora of programs with increasing feature entropy.

\section{Experiments}

We now empirically evaluate \systemname{} in two key dimensions: (1) its ability to generate diverse and complex synthetic verified program datasets across multiple languages and (2) the usefulness of the generated datasets for downstream tasks on existing programs. In particular, we answer the following two questions:

\begin{description}
        \item[RQ1: ]{Can \systemname{} generate diverse and complex synthetic verified programs at scale across multiple languages?}
        \item[RQ2: ]{Are generated programs useful for training models for verification-relevant downstream tasks?}
\end{description}

Furthermore, we also disentangle, via several ablations, the contributions of the \systemname{}'s multi-agent setup as well as the impact of using sampled documentation snippets and GitHub README files on the system's ability to generate diverse datasets.

\paragraph{Languages.} We instantiate \systemname{} for three verification-aware languages and tools: Dafny~\cite{leino2010dafny}, Verus (a tool for verification of Rust programs)~\cite{lattuada2023verus} and Frama-C~\cite{cuoq2012frama}. All three tools and languages are \emph{auto-active}: regular programs can be augmented by functional specifications (e.g., \emph{requires/ensures} clauses for functions), and proofs are generated automatically by invoking an SMT solver such as Z3~\cite{de2008z3}. Typically, complex specifications cannot be proved fully automatically from the implementation alone and require the program to also include user-written logical annotations, such as assertions and loop invariants, which help guide the SMT solver (e.g., by decomposing the problem into proving the annotations correct, and then assuming them to prove the verification conditions).

\paragraph{Models.} To implement workers in \systemname{}, we use Claude 4.5 Sonnet and Opus~\cite{anthropic2025opus45} as closed frontier models, and Qwen 2.5-Coder 32B as an open, dense code generation model~\cite{hui2024qwen2}. We use LoRA~\cite{hu2022lora} when fine-tuning Qwen 2.5-Coder 32B. All hyperparameters are listed in the appendix.

\subsection{Evaluating scalable synthetic data generation}

We first ask whether \systemname{} is effective at generating diverse verified programs at scale; Section~\ref{sec:downstream} will then assess whether our synthetic programs help improve models on downstream tasks related to verification. We start by analyzing whether \systemname{}'s workers are effective at generating programs by performing their own tasks (initiating, fixing and extending programs), and whether the system is effective even when using the open Qwen model, which can be cheaply applied at scale.

\begin{figure}
    \centering
    \includegraphics[width=0.9\linewidth]{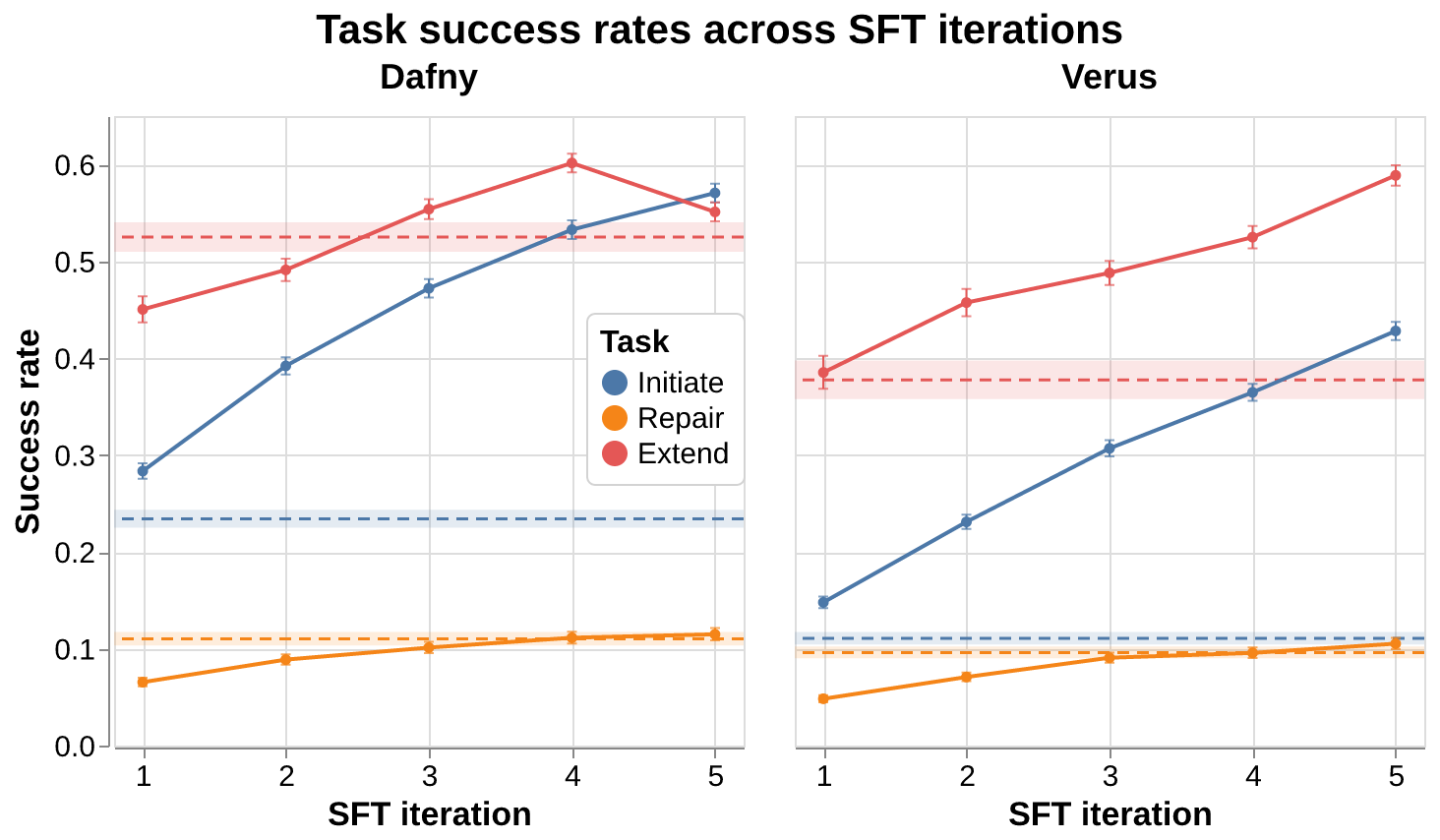}
    \caption{Task success rates for \systemname{} worker tasks backed by varying LLMs. Each color indicates a different worker task (Initiator, Fixer and Extender workers from Section ~\ref{sec:workers}). The dashed line shows performance for the seed Claude Sonnet 4.5 and Opus 4.5 workers; solid lines show the Qwen 2.5 Coder 32B model after fine-tuning either on the Claude distillation data (SFT iteration 1) or on the aggregate dataset generated from Claude models and its own previous iterations (SFT iterations $\geq 2$). SFT always follows the entropy maximization strategy, including on iteration 1. Overall, Qwen 2.5 Coder 32B becomes increasingly successful throughout iterations, sometimes far surpassing the initial success rate of Claude-based workers.}
    \label{fig:task-success-rates}
\end{figure}

Figure~\ref{fig:task-success-rates} shows worker success rates at each of the three \systemname{} generation tasks across Dafny and Verus. Here, the dashed horizontal line indicates worker success rates when Claude models are used for the workers. We find that Claude can initiate a new program successfully between 11\% (Verus) and 23\% (Dafny) of the time. Repair tends to be a harder task, whereas extending an already working programs tends to be easier, with success rates above 40\% in both languages. Importantly, we also find that our pipeline successfully distills and improves this generation capability into Qwen 2.5-Coder: the first iteration of supervised fine-tuning, essentially distilling Claude's responses into Qwen 2.5-Coder, already yields effective workers. After 5 iterations of self-improving, we find that Qwen 2.5-Coder either matches or outperforms the original Claude models across all worker tasks and languages, allowing synthetic data generation to be highly scalable at low cost.

\begin{figure}
    \centering
    \includegraphics[width=0.9\linewidth]{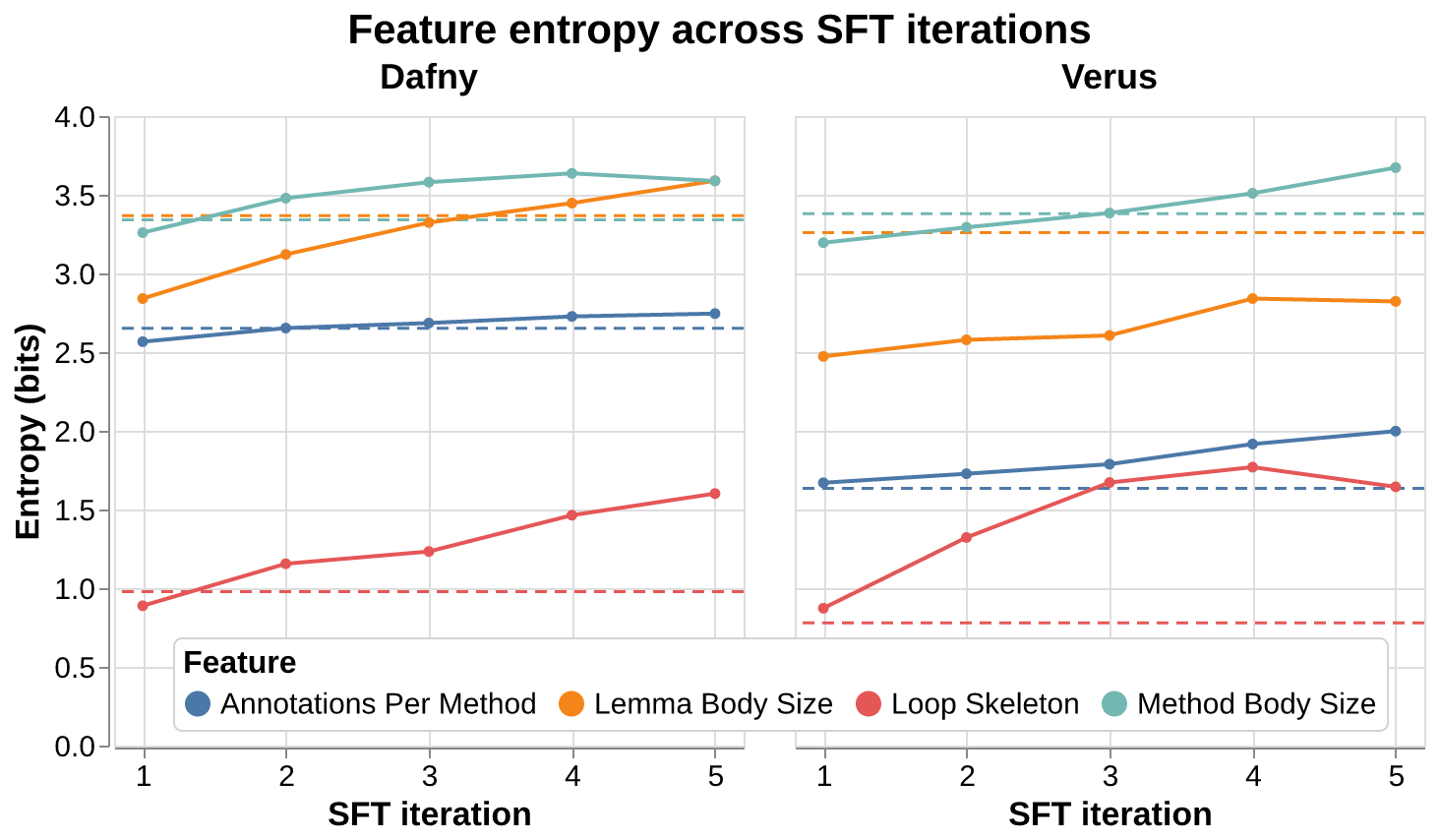}
    \caption{Feature entropy on the program corpora generated by \systemname{} across iterations, on the 4 features that entropy maximization targets: number of logical annotations per method, size of lemmas, method loop skeletons, and size of the bodies of methods. The dashed horizontal line shows the feature entropy on the corpus generated in the initial iteration with Claude 4.5 Sonnet and Opus; then, the solid line shows the entropy of that feature on each run using fine-tuned Qwen 2.5 Coder 32B on the combined corpus of examples generated so far, with ReST maximizing for entropy. Overall, \systemname{} effectively improves the diversity of the corpus over the chosen features across iterations.}
    \label{fig:feature-entropies}
\end{figure}

\subsection{Diversity}

We now evaluate the diversity of the synthetic datasets we generate both within our system and by comparing it to existing datasets --- both of human-written and synthetic programs --- across languages.

Figure~\ref{fig:feature-entropies} shows the feature entropy of the 4 features we optimize for (with the entropy maximization method from Section~\ref{sec:self-improvement}): the dashed line again shows the entropy of each feature in the corpus of programs generated from the initial runs with Claude, whereas the solid line shows how this entropy evolves in datasets generated after each round of SFT with Qwen when applying our entropy maximization method on the combined dataset of all programs generated so far. We observe that, across all features and languages, entropy tends to consistently increase across iterations. In fact, the last generated dataset with Qwen 2.5-Coder is more diverse, as measured by feature entropy, across all 4 features and languages, compared to the initial seed dataset generated by Claude Sonnet and Opus, with the only exception of lemma body sizes in Verus, where entropy increases as we run self-improvement iterations with Qwen but is still behind the Claude-generated datasets by iteration 5. Overall, this shows that \systemname{} can effectively \emph{increase diversity} over the chosen-features as it self-improves: not only workers succeed more often at their tasks, but also generate more spread-out feature distributions. This is in contrast with the conventional wisdom in training LLMs on their own self-generated data, where diversity collapse is a common concern --- one that our entropy maximization method effectively overcomes.

\textbf{Qualitative examples of increasing program complexity.} Beyond aggregate feature entropy, we also observe a qualitative increase in the complexity of individual programs across self-improvement iterations. Using the size (in lines) of the largest single method or lemma body in a program as a crude proxy for structural complexity --- directly one of our four entropy-maximization target features --- we compare the Claude seed corpus against the last Qwen self-improvement iteration for each language. We observe that median and 90th-percentile sizes (among many other distribution statistics) shift upward significantly in Dafny and Frama-C (e.g., from a median of 13 to 21 lines for Dafny; see Table~\ref{tab:complexity-shift}), and more strikingly, the largest verified example produced by the last Qwen iteration exceeds the largest example anywhere in the Claude seed corpus by 22--62\% across all three languages (154 vs.\ 95 lines for Dafny, 105 vs.\ 86 for Verus, 107 vs.\ 75 for Frama-C; Table~\ref{tab:complexity-shift-max}). We show all six programs in full in Appendix~\ref{app:qualitative-examples}, demonstrating programs from the Qwen-generated corpora that have no counterpart with comparable complexity in the Claude seed corpus.

\subsubsection{Comparison to existing datasets}

We now also compare diversity across a range of program features with existing datasets. Here, a major challenge in this comparison is that datasets can vary wildly in size: Verus-Bench, for instance, has less than 200 programs, whereas VeruSyn~\cite{di2026reducingcostsproofsynthesis} has over 1 million programs. To perform this analysis, we borrow the conceptual tool of \emph{rarefaction curves}, a commonly used tool in ecology to estimate the taxonomic diversity of a sample of species collected in a given region~\cite{raup1975taxonomic}. A rarefaction curve displays a sample diversity metric in the $y$ axis (e.g., number of distinct species observed, or entropy over label distributions) evolving as we grow our sample size on the $x$ axis. In particular, it allows us to visually compare differently sized datasets by showing how fast diversity grows as we sample from each underlying distribution.

\begin{figure}
    \centering
    \includegraphics[width=\linewidth]{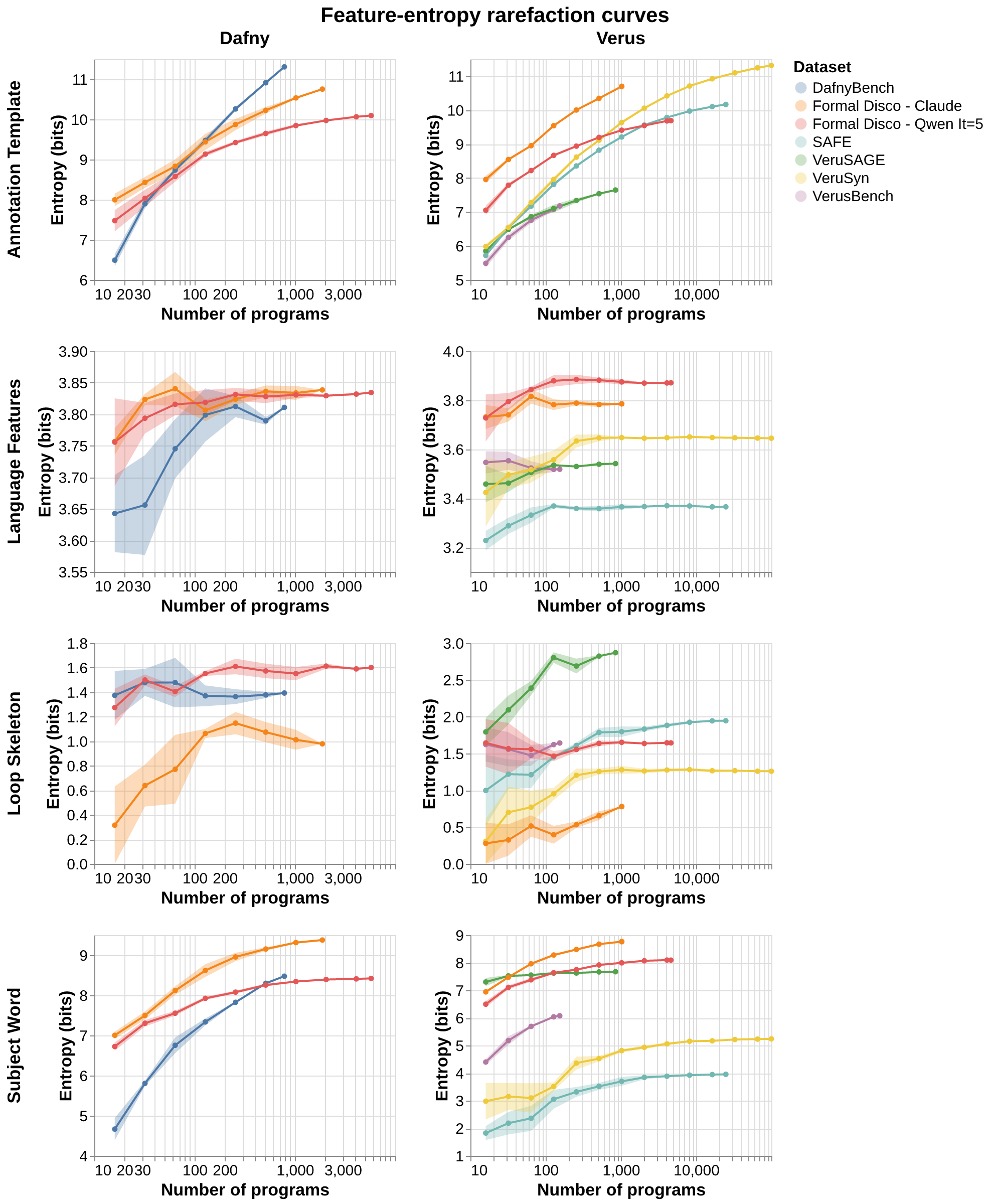}
    \caption{Rarefaction curves over 4 features, including three that we do not optimize for in our entropy maximization procedure (Language Features, Annotation Templates, Subject Words) and one that is optimized for (Loop Skeletons).}
    \label{fig:rarefaction-comparison-features1}
\end{figure}

Figure~\ref{fig:rarefaction-comparison-features2} shows rarefaction curves across 4 features, including three that we \emph{do not apply entropy maximization to} (language feature usage, annotation templates, and subject words) and one that we do (method loop skeletons). We make the following observations in each language:

\textbf{Dafny.} The human-written DafnyBench dataset has the most diverse logical annotations, whereas synthetic \systemname{} programs are more diverse on the other 3 features. Even though DafnyBench is considerably smaller than our synthetic datasets, the rarefaction curves over feature entropies over the syntactical skeleton of its logical annotations (e.g., assertions, loop invariants, requires/ensures clauses) shows that DafnyBench's human-written programs are still more diverse over these features. On this feature, which we do not optimize for using entropy maximization, Claude also remains more diverse than Qwen 2.5-Coder even after self-improvement (here we only show the last iteration of self-improvement for clarity; rarefaction curves comparing self-improvement iterations are shown later). In contrast, on the other 3 features (words in identifiers, loop skeletons and usage of Dafny features), we find that \systemname{}-generated programs are \emph{more diverse than DafnyBench}, even when matched for the same sample size. 

\textbf{Verus. } In Verus, we compare \systemname{}'s corpora with several existing datasets, including two synthetically-generated datasets from prior work (SAFE in cyan and VeruSyn in yellow). Here, we see that rarefaction curves of \systemname{}-generated datasets appear on top in 3 of the 4 features up to their respective lengths (annotation templates, language features, and subject words). In annotation templates, VeruSyn has not exhausted its diversity potential, and entropy keeps increasing (until eventually surpassing \systemname{}'s with Claude) even after sampling the entire dataset. In contrast, it saturates much earlier for the other three features. For loop skeletons, the most diverse dataset is VeruSAGE (green), consisting of Verus programs extracted from real-world repositories; here, SAFE also compares favorably to other datasets (despite, for instance, being used as the seed data in VeruSyn, which has less variation in loop structures). Overall, datasets show varied strengths in terms of their diversity, with synthetic datasets coming on top of real-world code across some features. Besides showing the competitiveness of \systemname{}-generated corpora, this analysis also illustrates rarefaction curves as a rich tool for comparing the distributions of heterogeneous datasets.

\subsection{Downstream tasks}
\label{sec:downstream}

Finally, we evaluate the performance of models trained on \systemname{}'s datasets on verification-related downstream tasks, allowing us to assess their usefulness beyond intrinsic success and diversity metrics.

\textbf{Logical annotation task. }
We first consider the logical annotation task as in prior work~\cite{poesia2024dafnyannotatoraiassistedverificationdafny}: given a program with formal specifications, insert logical annotations (i.e., assertions, loop invariants) so that the output program passes verification. For this task, we extract SFT training data from \systemname{}'s generated programs by simply stripping all annotations from the program and creating one SFT example that produces the diff that recovers the annotated program. We filter out trivial examples (those that verify even after removing annotations) and fine-tune Qwen 2.5-Coder 32B for 15k steps ($< 1$ epoch on the data) with LoRA on the $program \rightarrow diff$ examples (full hyperparameters can be found in the appendix). We do this separately for Dafny and Verus, and evaluate the resulting model on annotating DafnyBench~\cite{loughridge2024dafnybench} and VerusBench~\cite{yang2025autoverus} programs stripped of annotations and again filtering out programs that still verify. We are unaware of other large-scale datasets for fine-tuning on Dafny; for Verus, we also use our same data extraction pipeline to fine-tune the same model on the SAFE dataset~\cite{chen2026automatedproofgenerationrust}, containing 20k Verus programs generated by annotating real-world Rust programs. In both cases, we also compare to Claude 4.5 Opus with up to 4 independent attempts, prompted to do chain-of-thought reasoning before outputting its diff. Note that our fine-tuned models are not trained to produce a CoT, given the simple pipeline that extracts training data deterministically from programs, so our results with SFT should be seen as lower bounds on the performance that can be extracted from our datasets.

\begin{figure}
    \centering
    \includegraphics[width=0.8\linewidth]{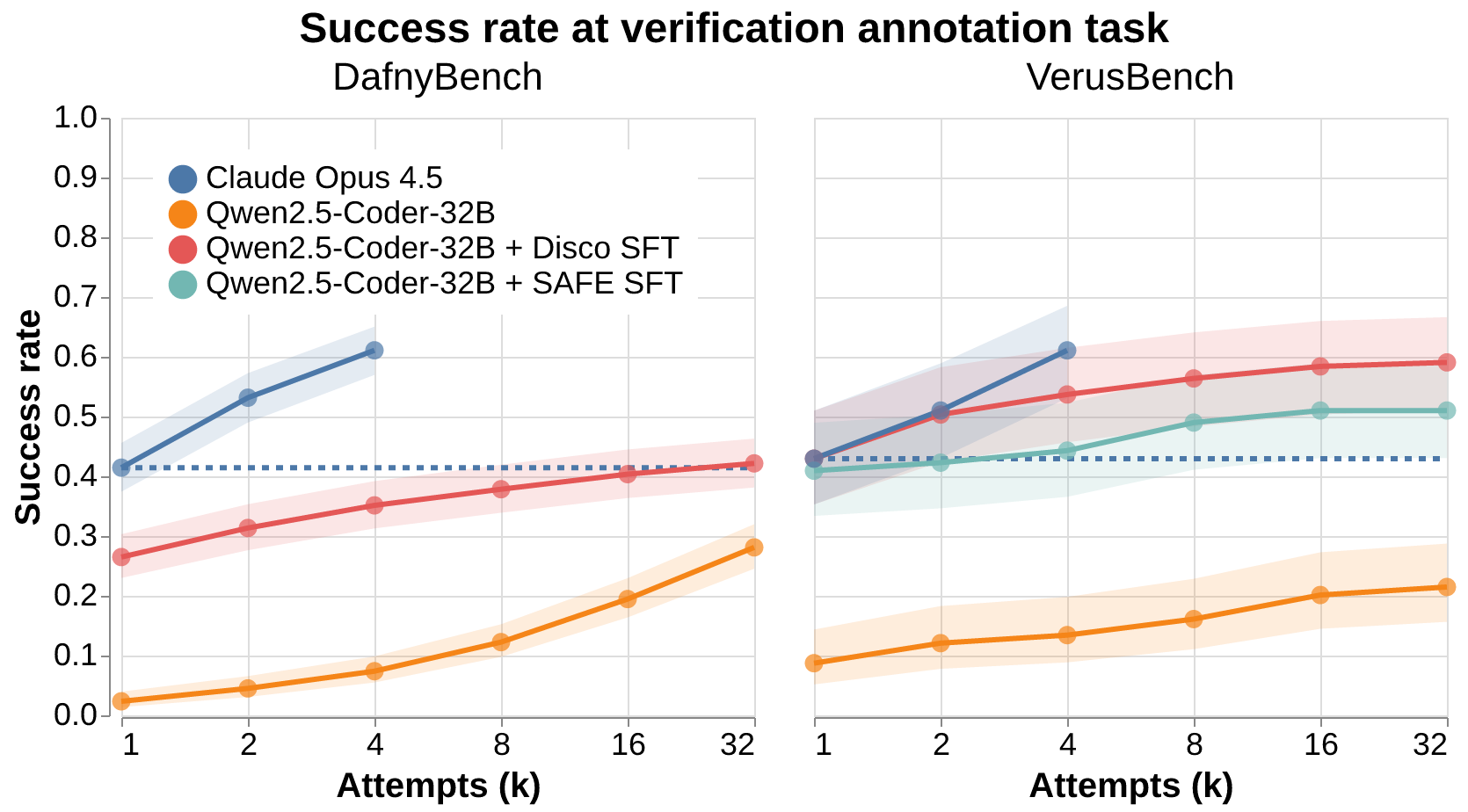}
    \caption{Success rate on the logical annotation task as a function of the
number of independent attempts $k$ (log scale), on DafnyBench (left; 555
programs) and VerusBench (right; 149 programs). Solid lines show Pass@$k$;
shaded bands are 95\% confidence intervals over benchmark
problems. The dashed line marks the single-attempt (Pass@1) success rate of
Claude 4.5 Opus, whose curve is shown up to $k=4$ attempts. Fine-tuning
Qwen 2.5-Coder 32B on data extracted from \systemname{}'s synthetic programs
roughly doubles Pass@16 on DafnyBench over the base model and matches
Claude 4.5 Opus's Pass@1 on VerusBench, where it also outperforms the same
model fine-tuned on SAFE~\cite{chen2026automatedproofgenerationrust}
data as the sampling budget grows.}
    \label{fig:fixer-pass-at-k}
\end{figure}

Figure~\ref{fig:fixer-pass-at-k} shows results for both languages in the verification annotation task. On DafnyBench, our fine-tuned Qwen 2.5-Coder at Pass@16 matches the single-shot performance of Claude 4.5 Opus ($40.4\%$ vs $41.4\%$); at this budget, the model trained on \systemname{} data is over 2x more successful than the base Qwen model (at $40.4\%\  vs\  19.5\%\ \text{Pass}@16$), showing significant gains from fine-tuning on our synthetic programs. On VerusBench, the Pass@1 performance of fine-tuned Qwen 2.5-Coder matches that of Opus 4.5, at exactly $43\%$, significantly ahead of the base Qwen 2.5-Coder 32B ($8.7\%$ success rate $\text{Pass}@1$). On VerusBench, a further point of comparison for the data quality obtained by \systemname{} is given by using our exact same SFT pipeline using the programs from SAFE, derived from real-world code: the SAFE-trained model is comparable at Pass@1 ($40.9\%$ vs $43.0\%$ for \systemname{} data), but scales worse with more attempts, plateauing at $51\%$ from Pass@16 onward while the \systemname{} model reaches $59.1\%$ at Pass@32 --- suggesting our synthetic programs yield more diverse repair attempts.

\begin{figure}
    \centering
    \includegraphics[width=0.45\linewidth]{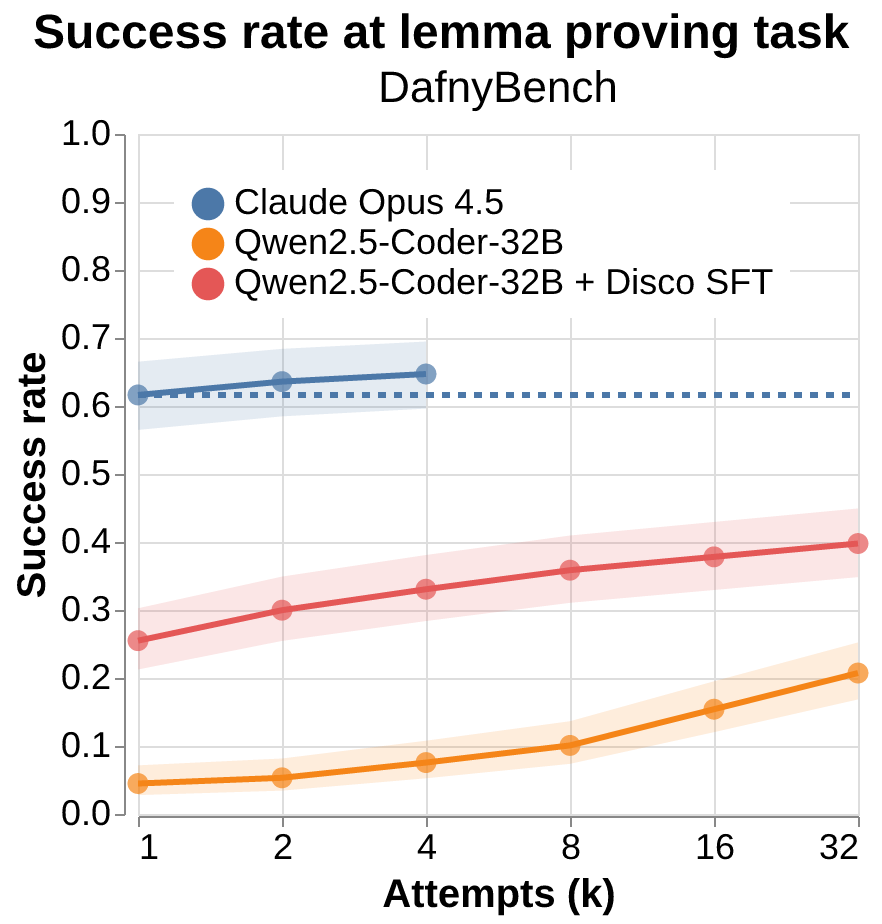}
    \caption{Success rate on the lemma proving task as a function of the number
of independent attempts $k$ (log scale), on 357 lemmas extracted from
DafnyBench (VerusBench contains no lemmas). Solid lines show Pass@$k$;
shaded bands are 95\% confidence intervals over benchmark
problems. The dashed line marks the Pass@1 success rate of Claude 4.5 Opus,
whose curve is shown up to $k=4$ attempts. A single sample from Qwen
2.5-Coder 32B fine-tuned on \systemname{} data outperforms 32 samples from
the base model; Claude 4.5 Opus performs best, but gains little from
additional attempts.}
    \label{fig:lemma-pass-at-k}
\end{figure}

\textbf{Lemma proving task. }
We next consider a lemma proving task: given a program containing a lemma
whose body (i.e., its proof) has been removed, generate a proof so that the
program verifies. As before, we extract SFT training data directly from
\systemname{}'s verified programs: we hollow out the body of each lemma and
create one SFT example that recovers the original proof, filtering out
trivial cases where the hollowed program still verifies (i.e., where Dafny
needs no manual proof). We fine-tune Qwen 2.5-Coder 32B on these
$\textit{program} \rightarrow \textit{proof}$ examples with the same setup as in the annotation
task, and evaluate on DafnyBench lemmas, hollowed and filtered in the same
way (357 lemmas in total); we only evaluate on Dafny here, since VerusBench
contains no lemmas. Figure~\ref{fig:lemma-pass-at-k} shows the results.
Fine-tuning on \systemname{} data improves Pass@1 over the base model by more
than $5\times$ ($25.5\%$ vs $4.5\%$); in fact, a single sample from the
fine-tuned model outperforms 32 samples from base Qwen 2.5-Coder ($20.7\%$
Pass@32), and the gap remains large at every sampling budget ($39.8\%$ vs
$20.7\%$ at Pass@32). Claude 4.5 Opus remains substantially stronger on this
task ($61.6\%$ Pass@1), consistent with proof generation benefiting from the
chain-of-thought reasoning that our fine-tuned models were not trained to
produce. Notably, however, the Opus 4.5 curve remains nearly flat with more attempts
(reaching only $64.7\%$ at Pass@4); our fine-tuned model continues to
improve steadily up to Pass@32.

Overall, these evaluations show that the fully
synthetic programs generated by \systemname{} are highly useful for improving
open models on verification tasks, matching a frontier model on Verus
annotation and yielding large gains across both languages. Moreover, since
SFT on deterministically extracted examples is arguably the simplest possible
way to use this data, our results should be read as lower bounds: richer
uses of the corpus --- e.g., distilling reasoning traces or reinforcement
learning against the verifier --- are promising directions for future work on making better use of large-scale corpora of examples for verification-aware languages.

\subsection{Ablations}
\label{sec:ablations}

Finally, we disentangle the contributions of three design choices in \systemname{}: the division of labor into three specialized workers (as opposed to a single monolithic agent), the GitHub READMEs in the Initiator's seed context, and the documentation snippets sampled from the language reference. Each ablation removes one component and compares the resulting runs on yield (verified programs per LLM call, with a maximum possible of $1$ if all calls were successful in generating a verified program) and on the diversity dimension that the component was designed to contribute. Since ablated runs produce corpora of different sizes, all diversity comparisons use rarefaction curves, which compare corpora at matched sample sizes.

\textbf{Monolithic agent vs. specialized workers.}
We compare \systemname{}'s three workers against a single ``monolithic'' agent that must produce a complete verified program in a single context window, running both with Claude 4.5 Opus on Dafny. If the generation fails, the monolithic agent is given verified feedback (like the fixer agent) in a growing context and allowed up to 3 attempts in repairing it. The three-worker setup produces verified programs at a higher rate per LLM call ($29.8\%$ vs $23.0\%$), but the more notable difference is in the \emph{complexity} of what each setup can reach: programs from the monolithic agent average 85 lines of code and never exceed 128, whereas the repair-and-extend loop yields programs averaging 268 lines, with the top decile above 598 lines and the largest at over 3,000 lines --- verified programs of a size that one-shot generation never attains. This validates the design intuition from Section~\ref{sec:workers}: the Initiator can afford to be unambitious because complexity is added incrementally by the Extender, with the Fixer recovering programs that each step breaks.

\begin{figure}
    \centering
    \begin{minipage}[t]{0.48\linewidth}
        \includegraphics[width=\linewidth]{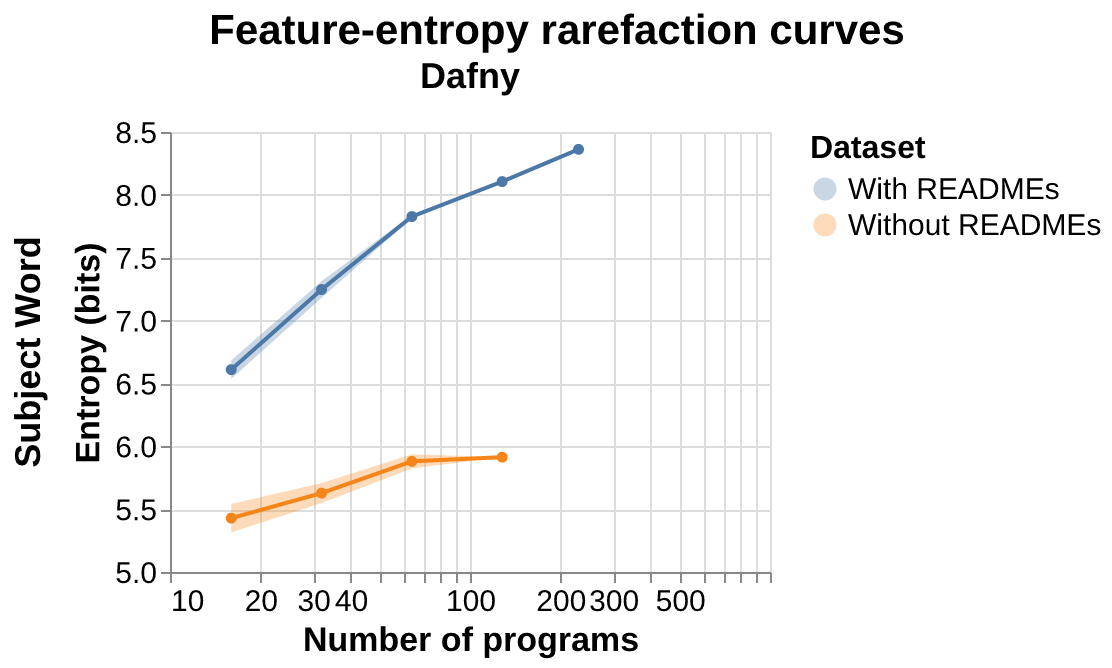}
    \end{minipage}\hfill
    \begin{minipage}[t]{0.48\linewidth}
        \includegraphics[width=\linewidth]{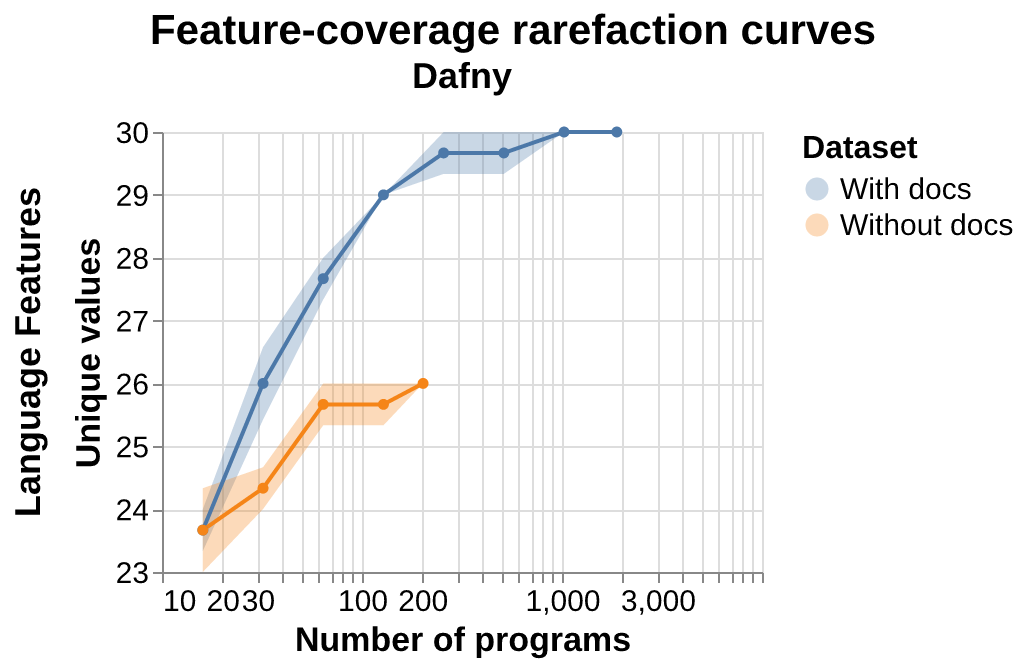}
    \end{minipage}
    \caption{Ablations of the two external entropy sources in the Initiator's seed context, on Dafny. \textbf{Left:} entropy over program subject words (themes) with and without GitHub READMEs (monolithic agent, Claude 4.5 Opus); without READMEs, generation highly concentrates onto common textbook data-structure themes and loses in entropy at every corpus size. \textbf{Right:} language-feature coverage (number of distinct tracked features observed, of 30) with and without documentation snippets (Claude 4.5 Opus and Sonnet runs pooled per condition); no-docs corpora plateau at 26 of 30 features, with four features never appearing at all. The fact that the orange curves are shorter also shows that the corresponding ablated agent produced less verified programs overall, even with the same LLM call budget.}
    \label{fig:ablations}
\end{figure}

\textbf{GitHub READMEs.}
We next remove the random GitHub READMEs from the seed context, letting the model invent program ideas on its own. Figure~\ref{fig:ablations} (left) shows rarefaction curves of the entropy over \emph{subject words} (content words extracted from identifiers, our proxy for program themes): without READMEs, theme entropy drops by over 2 bits at every sample size ($5.9$ vs $8.1$ bits at matched $N=128$; 151 vs 555 unique subject words). The collapse is visible to the naked eye: the most frequent subject words in the no-README corpus are \emph{stack, reverse, insert, pop, push, peek} and \emph{bst} --- the model gravitates to a handful of textbook data structures --- while the README-seeded corpus is spread over external themes (its most frequent subject word, \emph{cookie}, accounts for only $3.5\%$ of occurrences). Notably, READMEs also nearly \emph{double} the fraction of attempts yielding a verified program ($23.0\%$ vs $12.8\%$): concrete external themes appear to be easier to implement than self-invented ones, making READMEs beneficial even on raw yield.

\textbf{Documentation snippets.}
Last, we remove the documentation snippets from the seed context. Here the effect concentrates in the \emph{tail} of the language-feature distribution, which bulk entropy under-reports; Figure~\ref{fig:ablations} (right) therefore shows coverage rarefaction curves for Dafny, counting the number of distinct language features (of the 30 we track, derived from the language's reference) observed as the corpus grows. Corpora generated with documentation snippets quickly cover all $30/30$ features, while no-docs corpora plateau at $26/30$: \emph{bitvectors}, \emph{extreme predicates}, \emph{failure-compatible types} and \emph{iterators/coinduction} never appear at all without documentation snippets --- thus, we get zero entropy over those particular features. In entropy terms, both ablations confirm the hypotheses from Section~\ref{sec:workers}: READMEs and documentation snippets are complementary external entropy sources --- the former over program themes, the latter over language feature usage --- and removing either produces a measurable collapse precisely on the corresponding features.

\section{Conclusion}

We introduced \systemname{}, a system for open-ended generation of
formally verified programs, in which LLM-based workers collaborate through a
shared agenda to initiate, repair, and extend programs under compiler and
verifier feedback. Guiding both the design of the system and its
self-improvement loop is a principle of maximum entropy for synthetic data
generation, which unifies common intuitions about dataset quality --- including dataset size,
language feature coverage, program complexity --- into a single measurable
objective. Using \systemname{}, we generated and release large datasets of
verified programs in Dafny, Verus, and Frama-C --- in the case of Dafny and Frama-C, our datasets are larger than existing public corpora we are aware of by orders of magnitude. We showed that iterative
fine-tuning with entropy maximization yields workers that match or surpass the
frontier models they were distilled from, while producing \emph{increasingly}
diverse programs --- countering the diversity collapse commonly associated
with training on self-generated data. Fine-tuning open models on data
extracted from our corpora yields large gains on downstream verification
tasks on human-written programs.

Our work has several limitations that suggest directions for future work.
First, our downstream results likely understate the value of the generated
data: we train with the simplest possible recipe --- supervised fine-tuning on
examples extracted deterministically from programs, with no chain-of-thought
--- and the remaining gap to frontier models on reasoning-heavy tasks such as
lemma proving suggests substantial room for richer uses of the corpora,
such as distilling reasoning traces or reinforcement learning with
verifier feedback. Second, entropy maximization currently targets a small, hand-chosen
set of program features; on features we do not optimize for, such as the
structure of logical annotations, human-written programs remain more diverse
than our synthetic corpora, and automatically discovering \emph{which}
features to maximize entropy over (with the meta-level objective of improving downstream utility) is an open problem. Third, open-ended
generation still relies on external sources of entropy --- random READMEs and
documentation snippets --- without which we observe thematic collapse;
understanding how to sustain open-endedness with less external scaffolding
remains a challenge. Finally, while we instantiated \systemname{}
for three auto-active verification languages, the framework itself is
language-agnostic, and extending it to other formal settings --- including
interactive theorem provers and mathematics --- is a natural next step.

More broadly, our results suggest that the long-standing data barrier in
formal verification is surmountable: verifiers provide exactly the scalable
quality signal that synthetic data generation needs, and diversity can be
turned into a directly optimizable objective. As AI-generated
code becomes ubiquitous, we envision that new verification-aware languages
need not wait decades to accumulate the large-scale corpora that make AI assistance
effective --- they could launch with large synthetic datasets generated by
systems like \systemname{} from the first day, so that even their earliest adopters can program with
capable AI support, making strong correctness guarantees a practical default for software engineering in the era of AI coding agents.

\section*{Code and Data Availability}

\systemname{} is available at \url{https://github.com/metareflection/formal-disco}.

Our datasets generated from \systemname{} are available at \url{https://huggingface.co/collections/metareflection/formal-disco}.

\section*{Acknowledgments}
This work has been made possible in part by a gift from the Chan Zuckerberg Initiative Foundation to establish the Kempner Institute for the Study of Natural and Artificial Intelligence at Harvard University. Cluster experiments were partially made possible through a Kempner Institute Accelerator Award. This research was partially supported by a Coefficient Giving grant, funding and credits from an Amazon Research Award, and credits from Google Cloud. Simon Henniger was supported by the Harvard John A. Paulson School of Engineering and Applied Sciences (SEAS) Prize Fellowship and the German Academic Fellowship Organization, funded by the German Federal Ministry for Economic Affairs and Energy.

\bibliographystyle{plain}
\bibliography{references}

@article{gulcehre2023reinforced,
  title={Reinforced self-training (rest) for language modeling},
  author={Gulcehre, Caglar and Paine, Tom Le and Srinivasan, Srivatsan and Konyushkova, Ksenia and Weerts, Lotte and Sharma, Abhishek and Siddhant, Aditya and Ahern, Alex and Wang, Miaosen and Gu, Chenjie and others},
  journal={arXiv preprint arXiv:2308.08998},
  year={2023}
}

@article{dong2023raft,
  title={Raft: Reward ranked finetuning for generative foundation model alignment},
  author={Dong, Hanze and Xiong, Wei and Goyal, Deepanshu and Zhang, Yihan and Chow, Winnie and Pan, Rui and Diao, Shizhe and Zhang, Jipeng and Shum, Kashun and Zhang, Tong},
  journal={arXiv preprint arXiv:2304.06767},
  year={2023}
}

@inproceedings{leino2010dafny,
  title={Dafny: An automatic program verifier for functional correctness},
  author={Leino, K Rustan M},
  booktitle={International conference on logic for programming artificial intelligence and reasoning},
  pages={348--370},
  year={2010},
  organization={Springer}
}

@inproceedings{cuoq2012frama,
  title={{Frama-C}: A software analysis perspective},
  author={Cuoq, Pascal and Kirchner, Florent and Kosmatov, Nikolai and Prevosto, Virgile and Signoles, Julien and Yakobowski, Boris},
  booktitle={International conference on software engineering and formal methods},
  pages={233--247},
  year={2012},
  organization={Springer}
}

@inproceedings{lattuada2024verus,
  title={Verus: A practical foundation for systems verification},
  author={Lattuada, Andrea and Hance, Travis and Bosamiya, Jay and Brun, Matthias and Cho, Chanhee and LeBlanc, Hayley and Srinivasan, Pranav and Achermann, Reto and Chajed, Tej and Hawblitzel, Chris and others},
  booktitle={Proceedings of the ACM SIGOPS 30th Symposium on Operating Systems Principles},
  pages={438--454},
  year={2024}
}

@article{lattuada2023verus,
  title={Verus: Verifying rust programs using linear ghost types},
  author={Lattuada, Andrea and Hance, Travis and Cho, Chanhee and Brun, Matthias and Subasinghe, Isitha and Zhou, Yi and Howell, Jon and Parno, Bryan and Hawblitzel, Chris},
  journal={Proceedings of the ACM on Programming Languages},
  volume={7},
  number={OOPSLA1},
  pages={286--315},
  year={2023},
  publisher={ACM New York, NY, USA}
}

@inproceedings{de2008z3,
  title={Z3: An efficient {SMT} solver},
  author={De Moura, Leonardo and Bj{\o}rner, Nikolaj},
  booktitle={International conference on Tools and Algorithms for the Construction and Analysis of Systems},
  pages={337--340},
  year={2008},
  organization={Springer}
}

@techreport{anthropic2025opus45,
  title        = {Claude {Opus} 4.5 System Card},
  author       = {{Anthropic}},
  institution  = {Anthropic},
  year         = {2025},
  month        = nov,
  day          = {24},
  url          = {https://www-cdn.anthropic.com/bf10f64990cfda0ba858290be7b8cc6317685f47.pdf}
}

@article{hui2024qwen2,
  title={{Qwen2.5-Coder} Technical Report},
  author={Hui, Binyuan and Yang, Jian and Cui, Zeyu and Yang, Jiaxi and Liu, Dayiheng and Zhang, Lei and Liu, Tianyu and Zhang, Jiajun and Yu, Bowen and Lu, Keming and others},
  journal={arXiv preprint arXiv:2409.12186},
  year={2024}
}

@inproceedings{hu2022lora,
title={Lo{RA}: Low-Rank Adaptation of Large Language Models},
author={Edward J Hu and yelong shen and Phillip Wallis and Zeyuan Allen-Zhu and Yuanzhi Li and Shean Wang and Lu Wang and Weizhu Chen},
booktitle={International Conference on Learning Representations},
year={2022},
url={https://openreview.net/forum?id=nZeVKeeFYf9}
}

@article{lenat1984,
title = {Why am and eurisko appear to work},
journal = {Artificial Intelligence},
volume = {23},
number = {3},
pages = {269-294},
year = {1984},
issn = {0004-3702},
doi = {https://doi.org/10.1016/0004-3702(84)90016-X},
url = {https://www.sciencedirect.com/science/article/pii/000437028490016X},
author = {Douglas B. Lenat and John Seely Brown}
}

@misc{di2026reducingcostsproofsynthesis,
      title={Reducing the Costs of Proof Synthesis on Rust Systems by Scaling Up a Seed Training Set}, 
      author={Nongyu Di and Tianyu Chen and Shan Lu and Shuai Lu and Yeyun Gong and Peng Cheng and Jacob R. Lorch and Yuan Yao and Xiaoxing Ma},
      year={2026},
      eprint={2602.04910},
      archivePrefix={arXiv},
      primaryClass={cs.SE},
      url={https://arxiv.org/abs/2602.04910}, 
}

@misc{baksys2026atlasautomatedtoolkitlargescale,
      title={{ATLAS}: Automated Toolkit for Large-Scale Verified Code Synthesis}, 
      author={Mantas Baksys and Stefan Zetzsche and Olivier Bouissou and Remi Delmas and Soonho Kong and Sean B. Holden},
      year={2026},
      eprint={2512.10173},
      archivePrefix={arXiv},
      primaryClass={cs.SE},
      url={https://arxiv.org/abs/2512.10173}, 
}

@misc{chen2026automatedproofgenerationrust,
      title={Automated Proof Generation for Rust Code via Self-Evolution}, 
      author={Tianyu Chen and Shuai Lu and Shan Lu and Yeyun Gong and Chenyuan Yang and Xuheng Li and Md Rakib Hossain Misu and Hao Yu and Nan Duan and Peng Cheng and Fan Yang and Shuvendu K Lahiri and Tao Xie and Lidong Zhou},
      year={2026},
      eprint={2410.15756},
      archivePrefix={arXiv},
      primaryClass={cs.SE},
      url={https://arxiv.org/abs/2410.15756}, 
}

@misc{poesia2024dafnyannotatoraiassistedverificationdafny,
      title={dafny-annotator: {AI}-Assisted Verification of {Dafny} Programs}, 
      author={Gabriel Poesia and Chloe Loughridge and Nada Amin},
      year={2024},
      eprint={2411.15143},
      archivePrefix={arXiv},
      primaryClass={cs.SE},
      url={https://arxiv.org/abs/2411.15143}, 
}

@article{raup1975taxonomic,
  title={Taxonomic diversity estimation using rarefaction},
  author={Raup, David M},
  journal={Paleobiology},
  volume={1},
  number={4},
  pages={333--342},
  year={1975},
  publisher={Cambridge University Press}
}

@article{loughridge2024dafnybench,
  title={{DafnyBench}: A benchmark for formal software verification},
  author={Loughridge, Chloe and Sun, Qinyi and Ahrenbach, Seth and Cassano, Federico and Sun, Chuyue and Sheng, Ying and Mudide, Anish and Misu, Md Rakib Hossain and Amin, Nada and Tegmark, Max},
  journal={arXiv preprint arXiv:2406.08467},
  year={2024}
}

@article{yang2025autoverus,
  title={Autoverus: Automated proof generation for rust code},
  author={Yang, Chenyuan and Li, Xuheng and Misu, Md Rakib Hossain and Yao, Jianan and Cui, Weidong and Gong, Yeyun and Hawblitzel, Chris and Lahiri, Shuvendu and Lorch, Jacob R and Lu, Shuai and others},
  journal={Proceedings of the ACM on Programming Languages},
  volume={9},
  number={OOPSLA2},
  pages={3454--3482},
  year={2025},
  publisher={ACM New York, NY, USA}
}

@article{misu2024,
   title={Towards {AI}-Assisted Synthesis of Verified {Dafny} Methods},
   volume={1},
   ISSN={2994-970X},
   url={http://dx.doi.org/10.1145/3643763},
   DOI={10.1145/3643763},
   number={FSE},
   journal={Proceedings of the ACM on Software Engineering},
   publisher={Association for Computing Machinery (ACM)},
   author={Misu, Md Rakib Hossain and Lopes, Cristina V. and Ma, Iris and Noble, James},
   year={2024},
   month=jul,
pages={812–835} }

@inproceedings{sun2024clover,
author = {Sun, Chuyue and Sheng, Ying and Padon, Oded and Barrett, Clark},
title = {Clover: Closed-Loop Verifiable Code Generation},
year = {2024},
isbn = {978-3-031-65111-3},
publisher = {Springer-Verlag},
address = {Berlin, Heidelberg},
url = {https://doi.org/10.1007/978-3-031-65112-0_7},
doi = {10.1007/978-3-031-65112-0_7},
booktitle = {AI Verification: First International Symposium, SAIV 2024, Montreal, QC, Canada, July 22–23, 2024, Proceedings},
pages = {134–155},
numpages = {22},
location = {Montreal, QC, Canada}
}

@article{cassano2023multiple,
author = {Cassano, Federico and Gouwar, John and Nguyen, Daniel and Nguyen, Sydney and Phipps-Costin, Luna and Pinckney, Donald and Yee, Ming-Ho and Zi, Yangtian and Anderson, Carolyn Jane and Feldman, Molly Q and Guha, Arjun and Greenberg, Michael and Jangda, Abhinav},
title = {{MultiPL-E}: A Scalable and Polyglot Approach to Benchmarking Neural Code Generation},
year = {2023},
issue_date = {July 2023},
publisher = {IEEE Press},
volume = {49},
number = {7},
issn = {0098-5589},
url = {https://doi.org/10.1109/TSE.2023.3267446},
doi = {10.1109/TSE.2023.3267446},
journal = {IEEE Trans. Softw. Eng.},
month = jul,
pages = {3675–3691},
numpages = {17}
}

@article{cassano2024multiplt,
author = {Cassano, Federico and Gouwar, John and Lucchetti, Francesca and Schlesinger, Claire and Freeman, Anders and Anderson, Carolyn Jane and Feldman, Molly Q and Greenberg, Michael and Jangda, Abhinav and Guha, Arjun},
title = {Knowledge Transfer from High-Resource to Low-Resource Programming Languages for Code LLMs},
year = {2024},
issue_date = {October 2024},
publisher = {Association for Computing Machinery},
address = {New York, NY, USA},
volume = {8},
number = {OOPSLA2},
url = {https://doi.org/10.1145/3689735},
doi = {10.1145/3689735},
journal = {Proc. ACM Program. Lang.},
month = oct,
articleno = {295},
numpages = {32}
}

@article{robbes2026agentic,
  title={Agentic much? adoption of coding agents on GitHub},
  author={Robbes, Romain and Matricon, Th{\'e}o and Degueule, Thomas and Hora, Andre and Zacchiroli, Stefano},
  journal={ACM Transactions on Software Engineering and Methodology},
  year={2026},
  publisher={ACM New York, NY}
}

@misc{anthropic2026buildingccompiler,
  author       = {{Nicholas Carlini}},
  title        = {Building a C Compiler with a Team of Parallel Claudes},
  year         = {2026},
  month        = feb,
  day          = {5},
  howpublished = {\url{https://www.anthropic.com/engineering/building-c-compiler}}
}

@article{mugnier2025impact,
  title={On the Impact of Formal Verification on Software Development},
  author={Mugnier, Eric and Zhou, Yuanyuan and Jhala, Ranjit and Coblenz, Michael},
  journal={Proceedings of the ACM on Programming Languages},
  volume={9},
  number={OOPSLA2},
  pages={3642--3668},
  year={2025},
  publisher={ACM New York, NY, USA}
}

@inproceedings{gupta2021data,
  title={Data quality for machine learning tasks},
  author={Gupta, Nitin and Mujumdar, Shashank and Patel, Hima and Masuda, Satoshi and Panwar, Naveen and Bandyopadhyay, Sambaran and Mehta, Sameep and Guttula, Shanmukha and Afzal, Shazia and Sharma Mittal, Ruhi and others},
  booktitle={Proceedings of the 27th ACM SIGKDD conference on knowledge discovery \& data mining},
  pages={4040--4041},
  year={2021}
}

@article{herel2024collapse,
  title={Collapse of self-trained language models},
  author={Herel, David and Mikolov, Tomas},
  journal={arXiv preprint arXiv:2404.02305},
  year={2024}
}

@article{poesia2024learning,
  title={Learning formal mathematics from intrinsic motivation},
  author={Poesia, Gabriel and Broman, David and Haber, Nick and Goodman, Noah D},
  journal={Advances in Neural Information Processing Systems},
  volume={37},
  pages={43032--43057},
  year={2024}
}

@article{dong2025stp,
  title={Stp: Self-play llm theorem provers with iterative conjecturing and proving},
  author={Dong, Kefan and Ma, Tengyu},
  journal={arXiv preprint arXiv:2502.00212},
  year={2025}
}

@article{trinh2024solving,
  title={Solving olympiad geometry without human demonstrations},
  author={Trinh, Trieu H and Wu, Yuhuai and Le, Quoc V and He, He and Luong, Thang},
  journal={Nature},
  volume={625},
  number={7995},
  pages={476--482},
  year={2024},
  publisher={Nature Publishing Group UK London}
}

@article{xin2024deepseek,
  title={Deepseek-prover: Advancing theorem proving in llms through large-scale synthetic data},
  author={Xin, Huajian and Guo, Daya and Shao, Zhihong and Ren, Zhizhou and Zhu, Qihao and Liu, Bo and Ruan, Chong and Li, Wenda and Liang, Xiaodan},
  journal={arXiv preprint arXiv:2405.14333},
  year={2024}
}

@article{lin2025goedel,
  title={Goedel-prover: A frontier model for open-source automated theorem proving},
  author={Lin, Yong and Tang, Shange and Lyu, Bohan and Wu, Jiayun and Lin, Hongzhou and Yang, Kaiyu and Li, Jia and Xia, Mengzhou and Chen, Danqi and Arora, Sanjeev and others},
  journal={arXiv preprint arXiv:2502.07640},
  year={2025}
}

@article{wang2023voyager,
  title={Voyager: An open-ended embodied agent with large language models},
  author={Wang, Guanzhi and Xie, Yuqi and Jiang, Yunfan and Mandlekar, Ajay and Xiao, Chaowei and Zhu, Yuke and Fan, Linxi and Anandkumar, Anima},
  journal={arXiv preprint arXiv:2305.16291},
  year={2023}
}

@article{eldan2023tinystories,
  title={Tinystories: How small can language models be and still speak coherent english?},
  author={Eldan, Ronen and Li, Yuanzhi},
  journal={arXiv preprint arXiv:2305.07759},
  year={2023}
}

@article{li2023textbooks,
  title={Textbooks are all you need ii: phi-1.5 technical report},
  author={Li, Yuanzhi and Bubeck, S{\'e}bastien and Eldan, Ronen and Del Giorno, Allie and Gunasekar, Suriya and Lee, Yin Tat},
  journal={arXiv preprint arXiv:2309.05463},
  year={2023}
}

@article{shumailov2024ai,
  title={AI models collapse when trained on recursively generated data},
  author={Shumailov, Ilia and Shumaylov, Zakhar and Zhao, Yiren and Papernot, Nicolas and Anderson, Ross and Gal, Yarin},
  journal={Nature},
  volume={631},
  number={8022},
  pages={755--759},
  year={2024},
  publisher={Nature Publishing Group UK London}
}

\appendix

\section{Hyperparameters}
\label{app:hyperparameters}

We report the hyperparameters used for (1) synthetic data generation runs, (2) LoRA fine-tuning
during self-improvement and for the downstream tasks, and (3) evaluation.

\paragraph{Generation runs.} Each run schedules three workers (Initiator, Fixer, Extender) in
round robin. Table~\ref{tab:hparam-generation}
lists the settings shared across languages (Dafny, Verus, Frama-C). Runs with Qwen 2.5-Coder models took approximately $8-12h$ on a machine with $4x$ NVIDIA H100 GPUs (80 GB VRAM), served to the workers via vLLM. Workers were implemented using LangChain, which allowed accessing both Claude models (on AWS Bedrock) and the vLLM server (serving with OpenAI-compatible API) in a provider-agnostic manner. Each run used 48 parallel worker processes, all of them running the three workers in round-robin.

\begin{table}[ht]
\centering
\caption{Generation run hyperparameters.}
\label{tab:hparam-generation}
\begin{tabular}{@{}p{0.46\linewidth}p{0.46\linewidth}@{}}
\toprule
Parameter & Value \\
\midrule
Task attempts per run ($K$) & 10k with Claude models, 30k with Qwen \\
Reference doc snippets sampled per Initiator call & Uniform between 0--2 \\
Max repair attempts before a task is marked failed & 3 \\
LLM max output tokens & 10{,}000 \\
Seed frontier models & Claude Opus 4.5, Claude Sonnet 4.5 \\
Self-improvement backbone model & Qwen2.5-Coder-32B-Instruct \\
\bottomrule
\end{tabular}
\end{table}

Sampling temperature and top-$p$ for generation calls were left at each provider's default (not
explicitly pinned in configuration), for both frontier and open-weights models.

\paragraph{LoRA fine-tuning.} We use LoRA~\citep{hu2022lora} throughout, both for self-improvement
iterations on Qwen2.5-Coder-32B-Instruct and for the downstream fixer/lemma-proving tasks.
Table~\ref{tab:hparam-lora} lists the shared LoRA and optimization hyperparameters.

\begin{table}[h]
\centering
\caption{LoRA / SFT hyperparameters.}
\label{tab:hparam-lora}
\begin{tabular}{@{}p{0.46\linewidth}p{0.46\linewidth}@{}}
\toprule
Parameter & Value \\
\midrule
Base model & Qwen2.5-Coder-32B-Instruct \\
LoRA rank $r$ & 32 \\
LoRA $\alpha$ & 64 \\
LoRA dropout & 0.05 \\
Max sequence length & 8192 tokens (longer examples are discarded) \\
Learning rate & $2 \times 10^{-4}$ \\
LR scheduler & linear, warmup ratio 0.02 \\
Max gradient norm (gradient clipping) & 1.0 \\
Gradient steps & 15{,}000 ($< 1$ epoch) \\
Entropy-maximization filtering (self-improvement) & top $1/3$ of successful examples in the accumulated corpus so far, ranked by minimum
surprisal rank (mrr), stratified by prompt type (initiate/repair/extend) so that filtering does not change the base proportions of prompt types. \\
\bottomrule
\end{tabular}
\end{table}

\paragraph{Evaluation.} We evaluate downstream models with pass@$k$ for
$k \in \{1, 2, 4, 8, 16, 32\}$. For the Claude Opus 4.5 baseline we cap sampling at $k=4$
independent attempts (with chain-of-thought reasoning before the diff) given its cost; open and
fine-tuned Qwen models are sampled up to $k=32$. We report pass@16 as the headline comparison
point across models in the main text.

\section{Qualitative Examples of Increasing Program Corpus Complexity}
\label{app:qualitative-examples}

To complement the aggregate feature-entropy results of Section~\ref{sec:downstream}, we give
complete, verified examples showing that self-improvement with entropy maximization pushes programs into a complexity
regime the Claude seed corpus never reaches. We use the size (in non-blank lines) of the single
largest method or lemma body in a program as a crude scalar proxy for structural complexity --- this is
directly one of the four features targeted by entropy maximization (Section~\ref{sec:self-improvement}).
For each language, we compare the Claude Opus/Sonnet seed corpus against the last available
self-improvement iteration with Qwen2.5-Coder-32B (Iteration 5 for Dafny and Verus;
iteration 2 for Frama-C, which had still completed fewer completed self-improvement rounds at the time of writing).

\begin{table}[ht]
\centering
\caption{Median and 90th-percentile size (non-blank lines) of the largest method/lemma body per
program, Claude seed corpus vs.\ last Qwen self-improvement iteration. Verus uses the largest
\emph{lemma} body rather than method body, since Verus method-body size is dominated by a small
number of very long Claude extension chains that are not representative of typical complexity
(see text).}
\label{tab:complexity-shift}
\begin{tabular}{@{}p{0.22\linewidth}p{0.16\linewidth}p{0.16\linewidth}p{0.16\linewidth}p{0.16\linewidth}@{}}
\toprule
 & \multicolumn{2}{c}{Claude seed} & \multicolumn{2}{c}{Qwen, last iteration} \\
Language & median & p90 & median & p90 \\
\midrule
Dafny (method) & 13 & 28 & 21 & 41 \\
Verus (lemma) & 5 & 23 & 3 & 25 \\
Frama-C (method) & 14 & 25 & 15 & 30 \\
\bottomrule
\end{tabular}
\end{table}

Beyond this distributional shift, we manually verified the single largest method or lemma body produced in each corpus. Table~\ref{tab:complexity-shift-max}
shows that on all three languages, the last Qwen iteration produces a verified example
substantially past the largest example anywhere in the Claude seed corpus.

\begin{table}[ht]
\centering
\caption{Longest method or lemma body (non-blank lines) in each synthetic \systemname{} corpus.}
\label{tab:complexity-shift-max}
\begin{tabular}{@{}lll@{}}
\toprule
Language & Claude seed (max) & Qwen, last iteration (max) \\
\midrule
Dafny & 95 lines & 154 lines \\
Verus & 86 lines & 105 lines \\
Frama-C & 75 lines & 107 lines \\
\bottomrule
\end{tabular}
\\[2pt]
{\footnotesize Named declarations, shown in full below: Dafny --- \texttt{UpdateStatus} vs.\
\texttt{TestPromiseUsage}; Verus --- \texttt{lemma\_insert\_maintains\_sorted} vs.\
\texttt{values\_produce\_distinct\_chars}; Frama-C --- \texttt{parse\_cookie\_pair} vs.\
\texttt{find\_cookie\_by\_index}.}
\end{table}

We show each pair below in full.

\subsubsection{Dafny}

\begin{codebox}{Dafny --- Claude seed corpus | largest verified method (95 lines)}
\begin{lstlisting}[style=codestyle]
  method UpdateStatus(recordID: nat, newStatus: TaxonomicStatus, newReference: Option<nat>) returns (success: bool)
    requires Valid()
    requires ReferencesValid()
    requires recordID < |records|
    modifies this
    ensures Valid()
    ensures success ==> ReferencesValid()
    ensures !success ==> records == old(records) && ReferencesValid()
    ensures success ==> |records| == old(|records|)
    ensures success ==> records[recordID].status == newStatus
    ensures success ==> records[recordID].acceptedNameUsageID == newReference
    ensures success ==> records[recordID].scientificName == old(records[recordID].scientificName)
    ensures success ==> forall i :: 0 <= i < |records| && i != recordID ==> records[i] == old(records[i])
  {
    var oldRecord := records[recordID];
    var hasReference := newReference.Some?;
    
    // Check if the status transition is valid
    if !ValidStatusTransition(oldRecord.status, newStatus, hasReference) {
      success := false;
      return;
    }
    
    // Additional validation based on new status
    if newStatus == Synonym && newReference.None? {
      // Synonyms must have a reference
      success := false;
      return;
    }
    
    if newStatus == Accepted && newReference.Some? {
      // Accepted names cannot have a reference
      success := false;
      return;
    }
    
    // If new status is Synonym, verify the reference points to an Accepted taxon
    if newStatus == Synonym {
      assert newReference.Some?;
      var refID := newReference.value;
      if refID >= |records| || records[refID].status != Accepted || refID == recordID {
        // Cannot create self-referencing synonym or reference non-Accepted taxon
        success := false;
        return;
      }
    }
    
    // If the old record was Accepted and is being changed to non-Accepted,
    // check if any synonyms reference it
    if oldRecord.status == Accepted && newStatus != Accepted {
      var idx := 0;
      while idx < |records|
        invariant 0 <= idx <= |records|
        invariant forall k :: 0 <= k < idx && records[k].status == Synonym ==>
          records[k].acceptedNameUsageID != Some(recordID)
      {
        if records[idx].status == Synonym && records[idx].acceptedNameUsageID == Some(recordID) {
          // Cannot change status of a referenced accepted name
          success := false;
          return;
        }
        idx := idx + 1;
      }
      // At this point, we know no synonym references recordID
      assert forall k :: 0 <= k < |records| && records[k].status == Synonym ==>
        records[k].acceptedNameUsageID != Some(recordID);
    }
    
    // Create updated record
    var updatedRecord := TaxonRecord(oldRecord.scientificName, newStatus, newReference);
    
    // Verify the updated record is valid
    assert ValidRecord(updatedRecord);
    
    // Update the records sequence
    records := records[recordID := updatedRecord];
    
    // Verify that other synonym references are still valid
    // Verify that all synonym references are still valid after the update
    ghost var i := 0;
    while i < |records|
      invariant 0 <= i <= |records|
      invariant forall k :: 0 <= k < i && records[k].status == Synonym ==>
        exists j :: 0 <= j < |records| &&
          records[k].acceptedNameUsageID == Some(j) &&
          records[j].status == Accepted
    {
      if records[i].status == Synonym {
        assert records[i].acceptedNameUsageID.Some?;
        var refID := records[i].acceptedNameUsageID.value;
        if i == recordID {
          // This is the record we just updated to Synonym
          // We already verified that refID points to an Accepted taxon
          assert newStatus == Synonym;
          assert newReference.Some?;
          assert refID == newReference.value;
          assert refID < |records|;
          // The record at refID either wasn't changed (if refID != recordID)
          // or it's the new record (if refID == recordID, but that's impossible
          // since we just set recordID to Synonym status)
          // We verified earlier that refID < |records| and records[refID].status == Accepted
          // Since we're updating recordID to Synonym, refID must be different
          if refID == recordID {
            // This would mean we're creating a self-referencing synonym
            // But we checked earlier that records[refID].status == Accepted
            // before the update, and we're changing recordID to Synonym
            assert old(records[recordID].status) == Accepted;
            assert newStatus == Synonym;
            assert false; // Contradiction: we can't reference ourselves
          }
          assert records[refID] == old(records[refID]);
          assert records[refID].status == Accepted;
        } else {
          // This is an old synonym record that wasn't changed
          assert records[i] == old(records[i]);
          assert old(records[i].acceptedNameUsageID) == Some(refID);
          assert old(ReferencesValid());
          assert exists j :: 0 <= j < old(|records|) &&
            old(records[i].acceptedNameUsageID) == Some(j) &&
            old(records[j].status) == Accepted;
          // The refID must have been valid before
          assert refID < |old(records)|;
          assert old(records[refID].status) == Accepted;
          
          if refID == recordID {
            // This synonym referenced the record we're updating
            // We already checked that if oldRecord was Accepted and we're changing it,
            // no synonyms reference it - so this case is impossible
            assert old(records[recordID].status) == Accepted;
            assert oldRecord.status == Accepted;
            // If we're here and oldRecord.status == Accepted and newStatus != Accepted,
            // then we should have found this earlier and returned false
            if newStatus != Accepted {
              // We checked earlier that no synonyms reference recordID in this case
              assert forall k :: 0 <= k < |records| && records[k].status == Synonym ==>
                records[k].acceptedNameUsageID != Some(recordID);
              assert records[i].status == Synonym;
              assert records[i].acceptedNameUsageID == Some(recordID);
              assert false;
            }
            // We returned false earlier if any synonym referenced recordID
            // If we're here, newStatus == Accepted, so the reference is still valid
            assert newStatus == Accepted;
            assert records[refID].status == Accepted;
          } else {
            // The refID points to a different record that wasn't changed
            assert records[refID] == old(records[refID]);
            assert old(records[refID].status) == Accepted;
            assert records[refID].status == Accepted;
          }
        }
      }
      i := i + 1;
    }
    
    success := true;
  }
\end{lstlisting}
\end{codebox}

\begin{codebox}{Dafny --- Qwen | last self-improvement iteration | largest verified method (154 lines)}
\begin{lstlisting}[style=codestyle]
method TestPromiseUsage()
{
  var p := new Promise<int>();
  assert p.IsPending();
  
  var s1 := p.Fulfill(42);
  assert s1;
  assert p.IsFulfilled();
  assert p.GetValue() == 42;
  
  // Trying to fulfill again should fail
  var s2 := p.Fulfill(100);
  assert !s2;
  assert p.GetValue() == 42; // Value unchanged
  
  // Trying to reject should also fail
  var s3 := p.Reject("error");
  assert !s3;
  assert p.IsFulfilled(); // Still fulfilled, not rejected
  
  // Create a new promise and test rejection
  var q := new Promise<string>();
  assert q.IsPending();
  
  var s4 := q.Reject("something went wrong");
  assert s4;
  assert q.IsRejected();
  assert q.GetError() == "something went wrong";
  
  // Cannot fulfill after rejecting
  var s5 := q.Fulfill("value");
  assert !s5;
  assert q.IsRejected();
  
  // Test Then chaining
  var p2 := new Promise<int>();
  var s6 := p2.Fulfill(10);
  assert s6;
  
  // Chain with increment function
  var incremented := p2.Then(x => x + 1, _ => -1);
  assert incremented.IsFulfilled();
  assert incremented.GetValue() == 11;
  
  // Chain with negation function
  var negated := p2.Then(x => -x, _ => -1);
  assert negated.IsFulfilled();
  assert negated.GetValue() == -10;
  
  // Test error propagation via Then
  var p3 := new Promise<int>();
  var s7 := p3.Reject("original error");
  assert s7;
  
  // OnReject echoes the error message as a value
  var caught := p3.Then(x => x, err => |err| as int);
  assert caught.IsFulfilled();
  assert caught.GetValue() == 14; // "original error" has 14 characters
  
  // Test Catch handling
  var p4 := new Promise<int>();
  var s8 := p4.Reject("catch me");
  assert s8;
  
  // Catch transforms rejection into fulfillment
  var caught2 := p4.Catch(err => 999);
  assert caught2.IsFulfilled();
  assert caught2.GetValue() == 999;
  
  // Test chaining: Then followed by Catch
  var p5 := new Promise<int>();
  var s9 := p5.Reject("chain error");
  assert s9;
  
  // First Then should propagate the rejection
  var thenChained := p5.Then(x => x + 1, err => |err| as int);
  assert thenChained.IsFulfilled();
  assert thenChained.GetValue() == 11; // "chain error" has 11 chars
  
  // Catch after Then
  var p6 := new Promise<int>();
  var s10 := p6.Reject("then catch");
  assert s10;
  
  var t1 := p6.Then(x => x, err => -1);
  assert t1.IsFulfilled();
  assert t1.GetValue() == -1;
  
  var c1 := t1.Catch(err => 0);
  assert c1.IsFulfilled();
  assert c1.GetValue() == -1; // Catch doesn't change already fulfilled promise
  
  // Test multiple chained transformations
  var p7 := new Promise<int>();
  var s11 := p7.Fulfill(5);
  assert s11;
  
  // Chain: x => x + 3, then x => x * 2, then x => x - 1
  var chain1 := p7.Then(x => x + 3, _ => 0);
  assert chain1.IsFulfilled();
  assert chain1.GetValue() == 8;
  
  var chain2 := chain1.Then(x => x * 2, _ => 0);
  assert chain2.IsFulfilled();
  assert chain2.GetValue() == 16;
  
  var chain3 := chain2.Then(x => x - 1, _ => 0);
  assert chain3.IsFulfilled();
  assert chain3.GetValue() == 15;
  
  // Test error recovery in chain
  var p8 := new Promise<int>();
  var s12 := p8.Reject("early failure");
  assert s12;
  
  // First Then propagates the rejection
  var t2 := p8.Then(x => x, err => 100);
  assert t2.IsFulfilled();
  assert t2.GetValue() == 100;
  
  // Second Then continues the chain
  var t3 := t2.Then(x => x + 50, err => 0);
  assert t3.IsFulfilled();
  assert t3.GetValue() == 150;
  
  // Test that chaining order matters
  var p9 := new Promise<int>();
  var s13 := p9.Fulfill(10);
  assert s13;
  
  // Chain: x => x * 2, then x => x + 5
  var intermediate := p9.Then(x => x * 2, _ => 0);
  var doubleThenAdd := intermediate.Then(x => x + 5, _ => 0);
  assert doubleThenAdd.IsFulfilled();
  assert doubleThenAdd.GetValue() == 25; // (10 * 2) + 5 = 25
  
  // Reverse order: x => x + 5, then x => x * 2
  var intermediate2 := p9.Then(x => x + 5, _ => 0);
  var addThenDouble := intermediate2.Then(x => x * 2, _ => 0);
  assert addThenDouble.IsFulfilled();
  assert addThenDouble.GetValue() == 30; // (10 + 5) * 2 = 30
  
  // Test that original promise remains unchanged after chaining
  var p10 := new Promise<int>();
  var s14 := p10.Fulfill(42);
  assert s14;
  
  var chained := p10.Then(x => x + 1, _ => 0);
  assert chained.IsFulfilled();
  assert chained.GetValue() == 43;
  
  // Original promise still has its original value
  assert p10.IsFulfilled();
  assert p10.GetValue() == 42;
  
  // Test that multiple chains from same promise see the same state
  var p11 := new Promise<int>();
  var s15 := p11.Reject("race condition");
  assert s15;
  
  var chainA := p11.Then(x => x, err => 1);
  var chainB := p11.Then(x => x, err => 2);
  
  assert chainA.IsFulfilled();
  assert chainA.GetValue() == 1;
  
  assert chainB.IsFulfilled();
  assert chainB.GetValue() == 2;
  
  // Both chains observed the rejection, but handled it differently
  assert p11.IsRejected();
  assert p11.GetError() == "race condition";
  
  // Test complex error handling with multiple catches
  var p12 := new Promise<int>();
  var s16 := p12.Reject("initial error");
  assert s16;
  
  // First catch converts to value
  var catch1 := p12.Catch(err => 10);
  assert catch1.IsFulfilled();
  assert catch1.GetValue() == 10;
  
  // Second catch on already fulfilled promise is identity-like
  var catch2 := catch1.Catch(err => 20);
  assert catch2.IsFulfilled();
  assert catch2.GetValue() == 10;
  
  // Test that Then preserves the promise type correctly
  var p13 := new Promise<string>();
  var s17 := p13.Fulfill("hello");
  assert s17;
  
  // Transform string to its length
  var lengthPromise := p13.Then(s => |s| as int, err => 0);
  assert lengthPromise.IsFulfilled();
  assert lengthPromise.GetValue() == 5;
  
  // Test that error messages are preserved through Then
  var p14 := new Promise<int>();
  var s18 := p14.Reject("critical failure");
  assert s18;
  
  // Then propagates the rejection
  var thenPropagated := p14.Then(x => x + 1, err => |err| as int);
  assert thenPropagated.IsFulfilled();
  assert thenPropagated.GetValue() == 16; // "critical failure" has 16 chars
  
  // Test that the original promise's error is accessible
  assert p14.IsRejected();
  assert p14.GetError() == "critical failure";
  
  // Verify the promise state machine is deterministic
  // Pending -> Fulfilled -> (stays Fulfilled)
  var p15 := new Promise<int>();
  assert p15.IsPending();
  var s19 := p15.Fulfill(1);
  assert s19;
  assert p15.IsFulfilled();
  var s20 := p15.Fulfill(2);
  assert !s20; // Failed to fulfill again
  assert p15.GetValue() == 1; // Original value preserved
  
  // Pending -> Rejected -> (stays Rejected)
  var p16 := new Promise<int>();
  assert p16.IsPending();
  var s21 := p16.Reject("first");
  assert s21;
  assert p16.IsRejected();
  var s22 := p16.Reject("second");
  assert !s22; // Failed to reject again
  assert p16.GetError() == "first"; // Original error preserved
  
  // Verify that Once a promise is resolved, its state never changes
  // This is a fundamental property of promises
  var p17 := new Promise<int>();
  var s23 := p17.Fulfill(42);
  assert s23;
  assert p17.IsFulfilled();
  var s24 := p17.Resolve(Some("late error"), 100);
  assert !s24; // Cannot resolve after fulfillment
  assert p17.GetValue() == 42; // Value unchanged
}
\end{lstlisting}
\end{codebox}

\subsubsection{Verus}

\begin{codebox}{Verus --- Claude seed corpus | largest verified lemma (86 lines)}
\begin{lstlisting}[style=codestyle]
proof fn lemma_insert_maintains_sorted(jobs: Seq<Job>, job: Job)
    requires
        is_sorted_by_priority(jobs),
    ensures
        is_sorted_by_priority(insert_at(jobs, find_insert_position(jobs, job.priority), job)),
    decreases jobs.len(),
{
    lemma_insert_position_valid(jobs, job.priority);
    let pos = find_insert_position(jobs, job.priority);
    let result = insert_at(jobs, pos, job);
    
    if jobs.len() == 0 {
        assert(result.len() == 1);
        assert(is_sorted_by_priority(result));
    } else if priority_value(job.priority) > priority_value(jobs[0].priority) {
        // Insert at beginning
        assert(pos == 0);
        assert forall|i: int, j: int|
            0 <= i < j < result.len()
        implies
            priority_value(#[trigger] result[i].priority) >= priority_value(#[trigger] result[j].priority)
        by {
            if i == 0 {
                // result[0] is job, result[j] is jobs[j-1]
                assert(result[0] == job);
                assert(result[j] == jobs[j - 1]);
                assert(priority_value(job.priority) > priority_value(jobs[0].priority));
                if j > 1 {
                    assert(priority_value(jobs[0].priority) >= priority_value(jobs[j - 1].priority));
                }
            } else {
                // Both from original jobs
                assert(result[i] == jobs[i - 1]);
                assert(result[j] == jobs[j - 1]);
            }
        }
    } else {
        // Recursive case
        let rest = jobs.subrange(1, jobs.len() as int);
        lemma_insert_maintains_sorted(rest, job);

// In this branch, priority_value(job.priority) <= priority_value(jobs[0].priority)
// So find_insert_position returns 1 + find_insert_position(rest, ...)
assert(!(priority_value(job.priority) > priority_value(jobs[0].priority)));
// Unfold the definition of find_insert_position for this case
assert(jobs.len() > 0);
assert(jobs.len() != 0);
// The definition of find_insert_position: since jobs.len() > 0 and 
// !(priority_value(job.priority) > priority_value(jobs[0].priority)),
// it returns 1 + find_insert_position(rest, job.priority)
assert(find_insert_position(jobs, job.priority) == 1 + find_insert_position(rest, job.priority));
let rest_pos = find_insert_position(rest, job.priority);
lemma_insert_position_valid(rest, job.priority);
assert(rest_pos >= 0);
assert(pos == 1 + rest_pos);
assert(pos >= 1);
let rest_result = insert_at(rest, rest_pos, job);

// Prove result[0] == jobs[0] by expanding insert_at definition
assert(result == jobs.subrange(0, pos).push(job).add(jobs.subrange(pos, jobs.len() as int)));
let prefix = jobs.subrange(0, pos);
assert(prefix.len() >= 1);
assert(prefix[0] == jobs[0]);
assert(prefix.push(job)[0] == prefix[0]);
assert(prefix.push(job).add(jobs.subrange(pos, jobs.len() as int))[0] == prefix[0]);

// Show that result == jobs[0] followed by rest_result
        assert forall|i: int, j: int|
            0 <= i < j < result.len()
        implies
            priority_value(#[trigger] result[i].priority) >= priority_value(#[trigger] result[j].priority)
        by {
            if i == 0 {
                // result[0] is jobs[0]
                assert(result[0] == jobs[0]);
                if j < pos {
                    // result[j] is jobs[j]
                    assert(result[j] == jobs[j]);
                } else if j == pos {
                    // result[j] is job
                    assert(result[j] == job);
                    assert(priority_value(job.priority) <= priority_value(jobs[0].priority));
                } else {
                    // result[j] is jobs[j-1]
                    assert(result[j] == jobs[j - 1]);
                }
            } else {
                // Map back to rest_result indices
                let ri = i - 1;
                let rj = j - 1;
                if i < pos && j < pos {
                    assert(result[i] == jobs[i]);
                    assert(result[j] == jobs[j]);
                } else if i < pos && j == pos {
                    assert(result[i] == jobs[i]);
                    assert(result[j] == job);
                    // jobs[i] comes before insertion point, so its priority >= job.priority
                    assert(rest_result[ri] == rest[ri]);
                    assert(rest[ri] == jobs[i]);
                } else if i < pos && j > pos {
                    assert(result[i] == jobs[i]);
                    assert(result[j] == jobs[j - 1]);
                } else if i == pos {
                    assert(result[i] == job);
                    assert(result[j] == jobs[j - 1]);
                    // rest_result is sorted, and job is at position rest_pos in rest_result
                    assert(rest_result[ri] == job);
                    assert(rest_result[rj] == rest[rj - 1]);
                    assert(is_sorted_by_priority(rest_result));
                } else {
                    // Both after insertion point
                    assert(result[i] == jobs[i - 1]);
                    assert(result[j] == jobs[j - 1]);
                }
            }
        }
    }
}
\end{lstlisting}
\end{codebox}

\begin{codebox}{Verus --- Qwen | last self-improvement iteration | largest verified lemma (105 lines)}
\begin{lstlisting}[style=codestyle]
proof fn values_produce_distinct_chars(v1: int, v2: int)
    requires 
        0 <= v1 < 64,
        0 <= v2 < 64,
        value_to_char_spec(v1) == value_to_char_spec(v2),
    ensures v1 == v2,
{
    // Case analysis based on which ranges v1 and v2 fall into
    // Since value_to_char_spec produces unique characters for each valid value,
    // we can prove that if two values map to the same character, they must be equal.
    
    // Both in A-Z range
    if v1 < 26 && v2 < 26 {
        let c1 = (('A' as u8 + v1 as u8) as char);
        let c2 = (('A' as u8 + v2 as u8) as char);
        assert(value_to_char_spec(v1) == c1);
        assert(value_to_char_spec(v2) == c2);
        if value_to_char_spec(v1) == value_to_char_spec(v2) {
            assert(c1 == c2);
            assert(('A' as u8 + v1 as u8) as char == ('A' as u8 + v2 as u8) as char);
            assert(v1 as u8 == v2 as u8);
            assert(v1 == v2);
        }
    }
    // Both in a-z range
    else if (v1 >= 26 && v1 < 52) && (v2 >= 26 && v2 < 52) {
        let c1 = (('a' as u8 + (v1 - 26) as u8) as char);
        let c2 = (('a' as u8 + (v2 - 26) as u8) as char);
        assert(value_to_char_spec(v1) == c1);
        assert(value_to_char_spec(v2) == c2);
        if value_to_char_spec(v1) == value_to_char_spec(v2) {
            assert(c1 == c2);
            assert(('a' as u8 + (v1 - 26) as u8) as char == ('a' as u8 + (v2 - 26) as u8) as char);
            assert((v1 - 26) as u8 == (v2 - 26) as u8);
            assert(v1 == v2);
        }
    }
    // Both in 0-9 range  
    else if (v1 >= 52 && v1 < 62) && (v2 >= 52 && v2 < 62) {
        let c1 = (('0' as u8 + (v1 - 52) as u8) as char);
        let c2 = (('0' as u8 + (v2 - 52) as u8) as char);
        assert(value_to_char_spec(v1) == c1);
        assert(value_to_char_spec(v2) == c2);
        if value_to_char_spec(v1) == value_to_char_spec(v2) {
            assert(c1 == c2);
            assert((v1 - 52) as u8 == (v2 - 52) as u8);
            assert(v1 == v2);
        }
    }
    // v1 is 62 ('-'), v2 is 62 or 63
    else if v1 == 62 {
        assert(value_to_char_spec(v1) == '-');
        if v2 == 62 {
            assert(value_to_char_spec(v2) == '-');
        } else if v2 == 63 {
            assert(value_to_char_spec(v2) == '_');
        }
    }
    // v1 is 63 ('_'), v2 is 62 or 63
    else if v1 == 63 {
        assert(value_to_char_spec(v1) == '_');
        if v2 == 62 {
            assert(value_to_char_spec(v2) == '-');
        } else if v2 == 63 {
            assert(value_to_char_spec(v2) == '_');
        }
    }
    // v1 in 0-9, v2 in A-Z, a-z, 62, 63
    else if v1 >= 52 && v1 < 62 {
        let c1 = (('0' as u8 + (v1 - 52) as u8) as char);
        assert(value_to_char_spec(v1) == c1);
        // Check v2 in A-Z
        if v2 < 26 {
            let c2 = (('A' as u8 + v2 as u8) as char);
            assert(value_to_char_spec(v2) == c2);
            // Digit != Uppercase letter
        }
        // Check v2 in a-z
        else if v2 < 52 {
            let c2 = (('a' as u8 + (v2 - 26) as u8) as char);
            assert(value_to_char_spec(v2) == c2);
            // Digit != Lowercase letter
        }
        // Check v2 in 62, 63
        else if v2 == 62 {
            assert(value_to_char_spec(v2) == '-');
        }
        else if v2 == 63 {
            assert(value_to_char_spec(v2) == '_');
        }
    }
    // v1 in a-z, v2 in A-Z, 0-9, 62, 63
    else if v1 >= 26 && v1 < 52 {
        let c1 = (('a' as u8 + (v1 - 26) as u8) as char);
        assert(value_to_char_spec(v1) == c1);
        // Check v2 in A-Z
        if v2 < 26 {
            let c2 = (('A' as u8 + v2 as u8) as char);
            assert(value_to_char_spec(v2) == c2);
            // Lowercase != Uppercase
        }
        // Check v2 in 0-9
        else if v2 >= 52 && v2 < 62 {
            let c2 = (('0' as u8 + (v2 - 52) as u8) as char);
            assert(value_to_char_spec(v2) == c2);
            // Lowercase != Digit
        }
        // Check v2 in 62, 63
        else if v2 == 62 {
            assert(value_to_char_spec(v2) == '-');
        }
        else if v2 == 63 {
            assert(value_to_char_spec(v2) == '_');
        }
    }
    // v1 in A-Z, v2 in 0-9, a-z, 62, 63
    else if v1 < 26 {
        let c1 = (('A' as u8 + v1 as u8) as char);
        assert(value_to_char_spec(v1) == c1);
        // Check v2 in 0-9
        if v2 >= 52 && v2 < 62 {
            let c2 = (('0' as u8 + (v2 - 52) as u8) as char);
            assert(value_to_char_spec(v2) == c2);
            // Uppercase != Digit
        }
        // Check v2 in a-z
        else if v2 >= 26 && v2 < 52 {
            let c2 = (('a' as u8 + (v2 - 26) as u8) as char);
            assert(value_to_char_spec(v2) == c2);
            // Uppercase != Lowercase
        }
        // Check v2 in 62, 63
        else if v2 == 62 {
            assert(value_to_char_spec(v2) == '-');
        }
        else if v2 == 63 {
            assert(value_to_char_spec(v2) == '_');
        }
    }
}
\end{lstlisting}
\end{codebox}

\subsubsection{Frama-C}

\begin{codebox}{Frama-C --- Claude seed corpus | largest verified function (75 lines)}
\begin{lstlisting}[style=codestyle]
/*@
  requires \valid(cookie_str + (0 .. len-1));
  requires len >= 0 && len < MAX_COOKIE_LEN;
  requires \valid(pair);
  assigns *pair;
  ensures \result == 0 || \result == -1;
  ensures \result == 0 ==> pair->valid == true;
  ensures \result == -1 ==> pair->valid == false;
  ensures pair->key_len < MAX_KEY_LEN;
  ensures pair->value_len < MAX_VALUE_LEN;
*/
int parse_cookie_pair(const char *cookie_str, size_t len, cookie_pair_t *pair) {
    pair->valid = false;
    pair->key_len = 0;
    pair->value_len = 0;
    
    
    /*@
      loop invariant 0 <= i <= MAX_KEY_LEN;
      loop assigns i, pair->key[0 .. MAX_KEY_LEN-1];
      loop variant MAX_KEY_LEN - i;
    */
    for (size_t i = 0; i < MAX_KEY_LEN; i++) {
        pair->key[i] = '\0';
    }
    
    /*@
      loop invariant 0 <= i <= MAX_VALUE_LEN;
      loop assigns i, pair->value[0 .. MAX_VALUE_LEN-1];
      loop variant MAX_VALUE_LEN - i;
    */
    for (size_t i = 0; i < MAX_VALUE_LEN; i++) {
        pair->value[i] = '\0';
    }
    
    if (len == 0) {
        return -1;
    }
    
    
    size_t eq_pos = len; 
    
    /*@
      loop invariant 0 <= i <= len;
      loop invariant eq_pos == len || eq_pos < len;
      loop invariant eq_pos < len ==> cookie_str[eq_pos] == '=';
      loop assigns i, eq_pos;
      loop variant len - i;
    */
    for (size_t i = 0; i < len; i++) {
        if (cookie_str[i] == '=') {
            eq_pos = i;
            break;
        }
    }
    
    
    if (eq_pos == len || eq_pos == 0) {
        return -1;
    }
    
    
    size_t key_len = eq_pos;
    if (key_len >= MAX_KEY_LEN) {
        return -1; 
    }
    
    /*@
      loop invariant 0 <= i <= key_len;
      loop invariant i <= MAX_KEY_LEN;
      loop assigns i, pair->key[0 .. MAX_KEY_LEN-1];
      loop variant key_len - i;
    */
    for (size_t i = 0; i < key_len; i++) {
        pair->key[i] = cookie_str[i];
    }
    pair->key[key_len] = '\0';
    pair->key_len = key_len;
    
    
    size_t value_start = eq_pos + 1;
    size_t value_len = 0;
    
    if (value_start < len) {
        value_len = len - value_start;
    }
    
    if (value_len >= MAX_VALUE_LEN) {
        return -1; 
    }
    
    /*@
      loop invariant 0 <= i <= value_len;
      loop invariant i <= MAX_VALUE_LEN;
      loop assigns i, pair->value[0 .. MAX_VALUE_LEN-1];
      loop variant value_len - i;
    */
    for (size_t i = 0; i < value_len; i++) {
        pair->value[i] = cookie_str[value_start + i];
    }
    pair->value[value_len] = '\0';
    pair->value_len = value_len;
    
    pair->valid = true;
    return 0;
}
\end{lstlisting}
\end{codebox}

\begin{codebox}{Frama-C --- Qwen | last self-improvement iteration | largest verified function (107 lines)}
\begin{lstlisting}[style=codestyle]
/*@
  requires \valid_read(cookie + (0 .. cookie_len-1));
  requires \valid(key_out + (0 .. MAX_KEY_LEN-1));
  requires \valid(value_out + (0 .. MAX_VALUE_LEN-1));
  requires cookie_len >= 0;
  requires index >= 0;
  requires \separated(cookie + (0 .. cookie_len-1), key_out + (0 .. MAX_KEY_LEN-1), value_out + (0 .. MAX_VALUE_LEN-1));
  assigns  key_out[0 .. MAX_KEY_LEN-1], value_out[0 .. MAX_VALUE_LEN-1];
  ensures  \result == 0 || \result == 1;
*/
int find_cookie_by_index(const char *cookie, size_t cookie_len,
                          size_t index,
                          char *key_out, char *value_out)
{
  size_t pos = 0;
  size_t current_index = 0;
  
  /*@
    loop invariant 0 <= pos <= cookie_len;
    loop invariant 0 <= current_index <= index;
    loop assigns   pos, current_index, key_out[0 .. MAX_KEY_LEN-1], value_out[0 .. MAX_VALUE_LEN-1];
    loop variant   cookie_len - pos + (index - current_index);
  */
  while (pos < cookie_len) {
    
    size_t start = pos;
    
    /*@
      loop invariant start <= pos <= cookie_len;
      loop invariant \forall integer j; start <= j < pos ==> cookie[j] == ' ';
      loop assigns   pos;
      loop variant   cookie_len - pos;
    */
    while (pos < cookie_len && cookie[pos] == ' ') {
      pos++;
    }
    
    if (pos >= cookie_len || cookie[pos] == ';') {
      break;
    }
    
    if (current_index == index) {
      
      size_t key_start = pos;
      
      /*@
        loop invariant pos <= key_start <= cookie_len;
        loop invariant \forall integer j; pos <= j < key_start ==> cookie[j] != '=';
        loop assigns   key_start;
        loop variant   cookie_len - key_start;
      */
      while (key_start < cookie_len && cookie[key_start] != '=') {
        key_start++;
      }
      
      if (key_start >= cookie_len) {
        key_out[0] = '\0';
        value_out[0] = '\0';
        return 0;
      }
      
      size_t key_len = key_start - pos;
      if (key_len >= MAX_KEY_LEN) {
        key_len = MAX_KEY_LEN - 1;
      }
      
      /*@
        loop invariant 0 <= i <= key_len;
        loop invariant key_len < MAX_KEY_LEN;
        loop assigns   i, key_out[0 .. key_len-1];
        loop variant   key_len - i;
      */
      for (size_t i = 0; i < key_len; i++) {
        key_out[i] = cookie[pos + i];
      }
      key_out[key_len] = '\0';
      
      
      size_t value_start = key_start + 1;
      
      if (value_start < cookie_len) {
        size_t value_end = value_start;
        
        /*@
          loop invariant value_start <= value_end <= cookie_len;
          loop invariant \forall integer j; value_start <= j < value_end ==> 
                         (cookie[j] != ';' && cookie[j] != '\0');
          loop assigns   value_end;
          loop variant   cookie_len - value_end;
        */
        while (value_end < cookie_len && cookie[value_end] != ';' && cookie[value_end] != '\0') {
          value_end++;
        }
        
        size_t value_len = value_end - value_start;
        if (value_len >= MAX_VALUE_LEN) {
          value_len = MAX_VALUE_LEN - 1;
        }
        
        /*@
          loop invariant 0 <= i <= value_len;
          loop invariant value_len < MAX_VALUE_LEN;
          loop assigns   i, value_out[0 .. value_len-1];
          loop variant   value_len - i;
        */
        for (size_t i = 0; i < value_len; i++) {
          value_out[i] = cookie[value_start + i];
        }
        value_out[value_len] = '\0';
        
        return 1;
      }
      
      key_out[0] = '\0';
      value_out[0] = '\0';
      return 0;
    }
    
    current_index++;
    
    
    size_t semi_colon = pos;
    
    /*@
      loop invariant pos <= semi_colon <= cookie_len;
      loop invariant \forall integer j; pos <= j < semi_colon ==> 
                     (cookie[j] != ';' && cookie[j] != '\0');
      loop assigns   semi_colon;
      loop variant   cookie_len - semi_colon;
    */
    while (semi_colon < cookie_len && cookie[semi_colon] != ';' && cookie[semi_colon] != '\0') {
      semi_colon++;
    }
    
    if (semi_colon < cookie_len && cookie[semi_colon] == ';') {
      pos = semi_colon + 1;
    } else {
      pos = semi_colon;
      break;
    }
  }
  
  key_out[0] = '\0';
  value_out[0] = '\0';
  return 0;
}
\end{lstlisting}
\end{codebox}

\section{Implementation Details}
\label{app:implementation}

We describe here implementation-level details of \systemname{} that complement the
system description in Section~\ref{sec:workers}.

\paragraph{Agenda tasks and priorities.} The distributed agenda exposes a small custom RPC
server with operations for workers to read and write tasks and objects (we also have a local agenda implementation for small / single-process runs). Concurrency in the agenda is controlled by a single
lock, held while claiming tasks or creating/updating objects; a worker claims the next available
task by walking a priority-sorted list and atomically marking it as being worked on. Objects
store a path, type, parent object paths, content, arbitrary properties, and an
\emph{interestingness} score; tasks store a type, parent object paths, and mutable status
(new/attempted/being-worked-on/done/failed) together with an attempt counter. The effective
priority of a task is its base priority multiplied by the interestingness of every object it
depends on, so tasks attached to more ``interesting'' objects. While we attempted to experiment with this ``interestingness'' factor, we did not end up relying on it much to influence the dynamics of the system. Thus, although this is currently one knob in the implementation that could be used to focus a run on programs that appeared to have rare or promising features (as opposed to the entropy maximization fine-tuning, which only takes place after a full run), 

\paragraph{Checkpointing.} The agenda checkpoints itself periodically based on the number of
RPC calls processed, serializing its full state (tasks, objects,
task outcomes, and attempt counters) atomically to disk. On restart, any task that was marked as
being worked on when the process stopped is reset to a retryable state, so an interrupted worker
never leaves a task permanently stuck. Agenda checkpoints are the primary output of the system: the final checkpoint is where we take both full verified examples (for downstream fine-tuning) as well as distillation examples (for the 3 worker tasks) used for self-improvement.

\paragraph{Verification pipeline.} Each language backend shares a common interface with three
possible verification outcomes: \textsc{Success}, \textsc{GoalUnproven} (the program type-checks
and the verifier runs, but some proof obligation is left unproven), and \textsc{Fail} (syntax or
other hard failures). Verification is performed by invoking the relevant compiler/verifier with a timeout, and
classifying the outcome from its standard output/error. The verifier's raw stdout and stderr are passed to the LLM verbatim as the
``notes'' in the ``repair'' prompts (Appendix~\ref{app:prompts}).

\paragraph{Scheduler.} The round-robin scheduler cycles through the configured workers
indefinitely, giving each one its fixed fuel budget per turn, until the agenda signals that the
run's maximum number of task attempts has been reached. A worker with no eligible task simply ends its turn early, leaving any unused fuel unspent; there is no explicit
signal exchanged between workers to skip a turn.

\paragraph{Distillation examples.} Every LLM call a worker makes is recorded as a distillation
example object with four core fields: the prompt type (e.g., \texttt{initiate}, \texttt{repair},
\texttt{extend}), the arguments used to build the prompt (e.g., sampled repository, README, and
doc snippets for \texttt{initiate}; the failing program and verifier notes for \texttt{repair}),
the raw model response, and the outcome of the resulting attempt. These examples are the basis for
both the initial distillation from frontier models and the entropy-maximization filtering used in
self-improvement (Section~\ref{sec:self-improvement}).

\section{Program Features}
\label{app:features}

Table~\ref{tab:features} lists the program features $f : \mathcal{P} \rightarrow \mathcal{V}_f$
used throughout our feature-entropy analysis (Section~\ref{sec:self-improvement}). The first four
features are the ones we directly target for entropy maximization during self-improvement; the
last three (language feature usage, annotation templates, and subject words) are used only for
comparing datasets in Section~\ref{sec:downstream} and are not optimized against.

\begin{table}[h]
\centering
\caption{Program features used for feature-entropy analysis. ``Optimized'' features are the
targets of entropy maximization during self-improvement (Section~\ref{sec:self-improvement}).}
\label{tab:features}
\begin{tabular}{@{}p{0.30\linewidth}p{0.12\linewidth}p{0.48\linewidth}@{}}
\toprule
Feature & Optimized? & One-line summary \\
\midrule
Annotations per method & Yes & \# of logical annotations attached to a method. \\
Lemma body size & Yes & \# of non-blank lines in a lemma's body. \\
Loop skeleton & Yes & syntactic shape of a method's (possibly nested) loops. \\
Method body size & Yes & \# of non-blank lines in a method's body. \\
Language feature usage & No & which language constructs a program uses. \\
Annotation template & No & syntactic skeleton of an annotation, identifiers abstracted. \\
Subject words & No & lemmatized nouns/verbs from declaration names. \\
\bottomrule
\end{tabular}
\end{table}

\paragraph{Features are multisets, not scalars, per program.} A single program almost always
contributes more than one observation to a feature: a program with several methods has one
loop skeleton and one annotation count \emph{per method}, a program with several lemmas has one
lemma size \emph{per lemma}, and a program's language-feature usage and identifiers are tallied
over every occurrence in the program. Concretely, our implementation computes each feature as a
\texttt{Counter} over the program (a multiset mapping each observed value to how many times it
occurs), rather than as a single scalar. This requires no change to the entropy-maximization
framework of Section~\ref{sec:self-improvement}: we simply pool every program's counter into a
single dataset-level counter before normalizing to obtain $P^{\mathcal{D}}_f$ in
Equation~\ref{eq:h}, so a program with more methods, lemmas, or annotations contributes
proportionally more mass to the pooled empirical distribution, much like how a longer
document contributes more $n$-grams to a corpus-level language model.

We illustrate all seven features on the same example Dafny program below, showing the exact
output of our feature extractor on it.

\begin{codebox}{Example program used throughout this section}
\begin{lstlisting}[style=codestyle]
// Returns the index of the maximum element of a nonempty array.
method FindMaxIndex(values: array<int>) returns (maxIndex: int)
  requires values.Length > 0
  ensures 0 <= maxIndex < values.Length
  ensures forall k :: 0 <= k < values.Length ==> values[k] <= values[maxIndex]
{
  maxIndex := 0;
  var i := 1;
  while i < values.Length
    invariant 1 <= i <= values.Length
    invariant 0 <= maxIndex < i
    invariant forall k :: 0 <= k < i ==> values[k] <= values[maxIndex]
  {
    if values[i] > values[maxIndex] {
      maxIndex := i;
    }
    i := i + 1;
  }
}

// Checks whether an array is sorted in non-decreasing order.
method IsSorted(values: array<int>) returns (result: bool)
  ensures result <==> forall a, b :: 0 <= a <= b < values.Length ==> values[a] <= values[b]
{
  result := true;
  var i := 1;
  while i < values.Length
    invariant 1 <= i <= values.Length
    invariant result <==> forall a, b :: 0 <= a <= b < i ==> values[a] <= values[b]
  {
    if values[i-1] > values[i] {
      result := false;
    }
    i := i + 1;
  }
}

// The maximum index found by FindMaxIndex is the unique such index.
lemma MaxIndexIsUnique(values: array<int>, i: int, j: int)
  requires values.Length > 0
  requires 0 <= i < values.Length && 0 <= j < values.Length
  requires forall k :: 0 <= k < values.Length ==> values[k] <= values[i]
  requires forall k :: 0 <= k < values.Length ==> values[k] <= values[j]
  requires values[i] != values[j]
  ensures false
{
  assert values[i] <= values[j];
  assert values[j] <= values[i];
}
\end{lstlisting}
\end{codebox}

\paragraph{Annotations per method.} Counts requires, ensures, invariant, assert, and decreases
clauses attached to a method (lemmas are excluded, see ``Lemma size'' below). On the example
program, this yields \emph{two} observations, one per method: \texttt{FindMaxIndex} has 6
(1 requires + 2 ensures + 3 invariants) and \texttt{IsSorted} has 3 (1 ensures + 2 invariants).
Extracted multiset: $\{6 \mapsto 1,\ 3 \mapsto 1\}$.

\paragraph{Lemma body size.} Non-blank lines in a lemma's body. The example program has a single
lemma, \texttt{MaxIndexIsUnique}, whose two-line body (the two \texttt{assert} statements) gives
the multiset $\{2 \mapsto 1\}$.

\paragraph{Loop skeleton.} The syntactic shape of a method's loops, with all non-loop code (e.g.,
the \texttt{if} statements guarding the updates) discarded. Both \texttt{FindMaxIndex} and
\texttt{IsSorted} contain a single, non-nested \texttt{while} loop with no other loops inside it,
so both reduce to the same skeleton and the example yields the multiset
$\{\texttt{"while \{ \}"} \mapsto 2\}$ --- illustrating that repeated values are common and are
exactly what the entropy of this feature is sensitive to.

\paragraph{Method body size.} Non-blank lines in a method's body, excluding its signature and
specification clauses. The example program's two methods give the multiset
$\{12 \mapsto 1,\ 11 \mapsto 1\}$ (\texttt{FindMaxIndex} and \texttt{IsSorted}, respectively).

\paragraph{Language feature usage.} For each of a curated set of language constructs (one short
description with a usage example per construct, the same descriptions sampled as reference
snippets into Initiator prompts, Appendix~\ref{app:prompts}), a regular expression counts how many
times the construct is used in the program, giving one count per detected construct. On the
example program, this yields a multiset with 9 distinct constructs, each with its own occurrence
count, from a single 40-line program: $\{\texttt{logical-ops} \mapsto 9,\ \texttt{lambdas-arrows}
\mapsto 8,\ \texttt{quantifiers} \mapsto 6,\ \texttt{numeric-types} \mapsto 6,\ \texttt{arrays}
\mapsto 3,\ \ldots\}$. We curated such constructs from each language's reference manual:  30 for Dafny, 30 for Verus, and 28 for
Frama-C.

\paragraph{Annotation template.} The syntactic skeleton of an assertion, invariant, requires, or
ensures clause, with every identifier-shaped token replaced by \texttt{*} (a fixed set of
language keywords and built-ins, e.g. \texttt{forall}, is left unabstracted). Even our small
three-declaration example program yields 13 distinct templates (each occurring once or twice),
e.g. $\{\texttt{"requires: *.* > 0"} \mapsto 2,\ \texttt{"assert: *[*] <= *[*]"} \mapsto 2,\
\texttt{"invariant: 0 <= * < *"} \mapsto 1,\ \ldots\}$ --- illustrating how quickly this feature's
support grows even on small programs, which is part of why it saturates slowly in the
rarefaction curves of Section~\ref{sec:downstream}.

\paragraph{Subject words.} Declaration names are split at \texttt{camelCase} boundaries, part of
speech-tagged, lemmatized, and filtered to nouns and non-stopword verbs. From the three
declaration names in the example program (\texttt{FindMaxIndex}, \texttt{IsSorted},
\texttt{MaxIndexIsUnique}), this yields the multiset $\{\texttt{index} \mapsto 2,\ \texttt{find}
\mapsto 1,\ \texttt{sort} \mapsto 1\}$: \texttt{max} and \texttt{unique} are dropped because the
part-of-speech tagger does not tag them as a noun or verb in context, and \texttt{is} is dropped
as a stopword.

\section{Prompts and Example Generations}
\label{app:prompts}

We give the exact prompts used by the Initiator worker (Section~\ref{sec:workers}) for each
language, and show one successful generation from the self-improved Qwen2.5-Coder-32B model for
each language, taken from the last self-improvement iteration. Fixer and Extender prompts follow
the same diff-based instructions across languages, differing only in the target language's name
and verifier; we show the Dafny repair prompt as a representative example.

\subsubsection{Initiator prompts}

\begin{promptbox}{Dafny --- Initiator system prompt}
\begin{lstlisting}[style=promptstyle]
You are an expert Dafny programmer. You will receive a GitHub repository name and its README. Your task is to:
1. Come up with a concise idea for a Dafny program inspired by the repository's theme. The repository is most likely unrelated to verified programming, so freely adapt or reinterpret its theme.
2. Immediately implement that idea as a self-contained Dafny program.

Your program should NOT try to implement the entire idea, which is likely to be overly ambitious to write in one go.
Start with e.g. a few functions at most, or a very basic class with only a couple of core methods, or prove a basic lemma, etc. You can also add comments on ideas to extend the program later.
The output must be valid Dafny code and compile/verify when possible. Keep this initial program concise.

Output ONLY Dafny source code. Include a brief comment at the top of the program describing the idea.
\end{lstlisting}
\end{promptbox}

\begin{promptbox}{Verus --- Initiator system prompt}
\begin{lstlisting}[style=promptstyle]
You are an expert Verus programmer. You will receive a GitHub repository name and its README. Your task is to:
1. Come up with a concise idea for a Verus program inspired by the repository's theme. The repository is most likely unrelated to verified programming, so freely adapt or reinterpret its theme.
2. Immediately implement that idea as a self-contained Verus program.

Your program should NOT try to implement the entire idea, which is likely to be overly ambitious to write in one go.
Start with e.g. a few functions at most, or prove a basic lemma, etc. You can also add comments on ideas to extend the program later.
The output must be valid Verus code (Rust with Verus verification annotations) and compile/verify when possible. Keep this initial program concise.

Output ONLY Verus source code. Include a brief comment at the top of the program describing the idea.
\end{lstlisting}
\end{promptbox}

\begin{promptbox}{Frama-C --- Initiator system prompt}
\begin{lstlisting}[style=promptstyle]
You are an expert C programmer who writes formally verified code using Frama-C and ACSL. You will receive a GitHub repository name and its README. Your task is to:
1. Come up with a concise idea for a C program with ACSL annotations inspired by the repository's theme. The repository is most likely unrelated to formal verification, so freely adapt or reinterpret its theme.
2. Immediately implement that idea as a self-contained C program with ACSL annotations.

Your program should NOT try to implement the entire idea, which is likely to be overly ambitious to write in one go.
Start with e.g. a few functions at most, or prove a basic lemma, etc. You can also add comments on ideas to extend the program later.
Include function contracts (requires, ensures, assigns) and loop annotations (loop invariant, loop assigns, loop variant) as appropriate.
The output must be valid C code with ACSL annotations and verify with frama-c -wp when possible. Keep this initial program concise.

Output ONLY C source code with ACSL annotations. Include a brief comment at the top describing the idea.
\end{lstlisting}
\end{promptbox}

The user message follows a shared template across languages, populated with a sampled repository
name, its README, and up to 2 sampled reference snippets (Section~\ref{sec:workers}):

\begin{promptbox}{Initiator user message template}
\begin{lstlisting}[style=promptstyle]
Repository: {repo}

README:
{readme}

[optional: up to 2 sampled reference snippets, see below]
Come up with an idea for a {LANGUAGE} program inspired by this repository and implement it. Output only {LANGUAGE} source code.
\end{lstlisting}
\end{promptbox}

where each reference snippet takes the form:

\begin{promptbox}{Reference snippet format}
\begin{lstlisting}[style=promptstyle]
Reference snippets for some {LANGUAGE} language constructs you may find useful:

--- {feature_id} ---
{feature description and one usage example, sampled from the language reference}
\end{lstlisting}
\end{promptbox}

\subsubsection{Repair prompt (representative)}

\begin{promptbox}{Dafny --- Fixer system prompt}
\begin{lstlisting}[style=promptstyle]
You are an expert Dafny developer. You will be given a Dafny program that has errors pointed out by Dafny. These errors can be syntactic, or failures to verify the program (i.e. prove post-conditions or verify current assertions/invariants).
Your job is to repair these errors by emitting a DIFF in a simple, line-based format.

Diff format:
- Lines starting with '@@' are anchors (search-forward markers). These don't modify the program, but just start a new 'block' of changes in your patch.
- Lines starting with '=' keep that exact line: find it forward and advance the cursor. You typically only need a few of these after your @@ line to position the cursor for the actual changes: you don't need to copy much of the original file.
- Lines starting with '-' delete that exact line found forward.
- Lines starting with '+' add a new line at the current cursor.

- All diff lines should start with one of the special characters above and a space following them. Other lines will be completely ignored
Here is an example of a diff:

Text before:
{example_before}

Example of model output (diff in the format you must follow):
{example_diff}

Text after:
{example_after}
\end{lstlisting}
\end{promptbox}

\subsubsection{Example generations}

We show one successful program generated by the self-improved Qwen2.5-Coder-32B model for each
language, together with the sampled README and reference snippet(s) that seeded it. All three
examples come from the Initiator task and pass compilation and verification.

\paragraph{Dafny.} Sampled repository: \texttt{TGITS/programming-workouts}.

\begin{promptbox}{Dafny example --- sampled README}
\begin{lstlisting}[style=promptstyle]
Repository: TGITS/programming-workouts

# Hello World

The classical introductory exercise. Just say "Hello, World!".

["Hello, World!"](http://en.wikipedia.org/wiki/%22Hello,_world!%22_program) is
the traditional first program for beginning programming in a new language
or environment.

The objectives are simple:

- Write a function that returns the string "Hello, World!".
- Run the test suite and make sure that it succeeds.
- Submit your solution and check it at the website.

If everything goes well, you will be ready to fetch your first real exercise.

## Running the tests

To run the tests, run the command `dotnet test` from within the exercise directory.

## Autoformatting the code

F# source code can be formatted with the [Fantomas](https://github.com/fsprojects/fantomas) tool.

After installing it with `dotnet tool restore`, run `dotnet fantomas .` to format code within the current directory.

## Further information

For more detailed information about the F# track, including how to get help if
you're having trouble, please visit the exercism.io [F# language page](http://exercism.io/languages/fsharp/resources).

## Source

This is an exercise to introduce users to using Exercism [http://en.wikipedia.org/wiki/%22Hello,_world!%22_program](http://en.wikipedia.org/wiki/%22Hello,_world!%22_program)
\end{lstlisting}
\end{promptbox}

\begin{promptbox}{Dafny example --- sampled reference snippet}
\begin{lstlisting}[style=promptstyle]
--- quantifiers ---
Quantifier expressions `forall x :: P(x)` and `exists x :: P(x)` are first-class boolean expressions. Bound variables may be constrained by a range (`forall x | C(x) :: P(x)`). Triggers (e.g., `{:trigger}` or matched terms appearing in P) guide the SMT solver in instantiating quantifiers.

```dafny
predicate AllPositive(s: seq<int>)
{
  forall i :: 0 <= i < |s| ==> s[i] > 0
}

predicate HasZero(s: seq<int>)
{
  exists i :: 0 <= i < |s| && s[i] == 0
}

lemma PositiveImpliesNoZero(s: seq<int>)
  requires AllPositive(s)
  ensures !HasZero(s)
{
}

lemma BoundedRange(n: int)
  requires n >= 0
  ensures forall k :: 0 <= k <= n ==> 0 <= k && k <= n
{
}

method UseQuantifiers()
{
  var s := [1, 2, 3];
  assert AllPositive(s);
  PositiveImpliesNoZero(s);
  assert !HasZero(s);
}
```
\end{lstlisting}
\end{promptbox}

\begin{codebox}{Dafny example --- Qwen2.5-Coder-32B response (verifies successfully)}
\begin{lstlisting}[style=codestyle]
// Inspired by "programming-workouts" - A classic "Hello, World!" exercise
// implemented in Dafny with verification. We define a function that returns
// "Hello, World!" and prove properties about string concatenation and 
// character sequences.

function HelloWorld(): string
{
  "Hello, World!"
}

// Prove that our HelloWorld function returns a string of expected length
lemma HelloWorldLength()
  ensures |HelloWorld()| == 13
{
}

// String concatenation utility
function Concat(s1: string, s2: string): string
{
  s1 + s2
}

// Prove that concatenation preserves lengths
lemma ConcatLength(s1: string, s2: string)
  ensures |Concat(s1, s2)| == |s1| + |s2|
{
}

// Prove that concatenating empty string doesn't change the string
lemma ConcatEmptyRight(s: string)
  ensures Concat(s, "") == s
{
}

lemma ConcatEmptyLeft(s: string)
  ensures Concat("", s) == s
{
}

// Extract greeting and audience from HelloWorld
lemma HelloWorldStructure()
  ensures HelloWorld() == "Hello" + ", World!"
  ensures |"Hello"| == 5
  ensures |", World!"| == 8
{
}

// A method that uses our HelloWorld function
method SayHello() returns (greeting: string)
  ensures greeting == "Hello, World!"
  ensures |greeting| == 13
{
  greeting := HelloWorld();
  HelloWorldLength();
}

// Future extensions could include:
// - More international greetings with locale parameters
// - ASCII art rendering of "Hello World"
// - Verification of different string operations (substring, replace, etc.)
// - Unicode character properties
\end{lstlisting}
\end{codebox}

\paragraph{Verus.} Sampled repository: \texttt{NirmalL/OWA}.

\begin{promptbox}{Verus example --- sampled README}
\begin{lstlisting}[style=promptstyle]
Repository: NirmalL/OWA

# cookie [![Build Status](https://secure.travis-ci.org/defunctzombie/node-cookie.png?branch=master)](http://travis-ci.org/defunctzombie/node-cookie) #

cookie is a basic cookie parser and serializer. It doesn't make assumptions about how you are going to deal with your cookies. It basically just provides a way to read and write the HTTP cookie headers.

See [RFC6265](http://tools.ietf.org/html/rfc6265) for details about the http header for cookies.

## how?

```
npm install cookie
```

```javascript
var cookie = require('cookie');

var hdr = cookie.serialize('foo', 'bar');
// hdr = 'foo=bar';

var cookies = cookie.parse('foo=bar; cat=meow; dog=ruff');
// cookies = { foo: 'bar', cat: 'meow', dog: 'ruff' };
```

## more

The serialize function takes a third parameter, an object, to set cookie options. See the RFC for valid values.

### path
> cookie path

### expires
> absolute expiration date for the cookie (Date object)

### maxAge
> relative max age of the cookie from when the client receives it (seconds)

### domain
> domain for the cookie

### secure
> true or false

### httpOnly
> true or false
\end{lstlisting}
\end{promptbox}

\begin{promptbox}{Verus example --- sampled reference snippet}
\begin{lstlisting}[style=promptstyle]
--- while-loops ---
A verified `while` loop carries `invariant` clauses that hold before the loop, after every iteration, and (with the negated guard) at the loop's exit. A `decreases` clause names a value that strictly decreases each iteration, proving termination.

```rust
use vstd::prelude::*;

verus! {

fn count_to(n: u32) -> (r: u32)
    requires n <= 1_000_000,
    ensures r == n,
{
    let mut i: u32 = 0;
    while i < n
        invariant
            i <= n,
            n <= 1_000_000,
        decreases n - i,
    {
        i = i + 1;
    }
    i
}

} // verus!
```
\end{lstlisting}
\end{promptbox}

\begin{codebox}{Verus example --- Qwen2.5-Coder-32B response (verifies successfully)}
\begin{lstlisting}[style=codestyle]
// Inspired by the cookie parsing library: A verified simple key-value pair parser
// for HTTP-style cookies (e.g., "key1=value1; key2=value2").
// We verify properties about parsing and serialization of cookie strings.

use vstd::prelude::*;
use vstd::string::*;

verus! {

// Represents a parsed cookie as a key-value pair
pub struct Cookie {
    pub key: Vec<u8>,
    pub value: Vec<u8>,
}

impl Cookie {
    pub open spec fn well_formed(&self) -> bool {
        &&& self.key.len() > 0
        &&& self.value.len() > 0
    }

    pub open spec fn key_len(&self) -> nat {
        self.key.len() as nat
    }

    pub open spec fn value_len(&self) -> nat {
        self.value.len() as nat
    }
}

// Spec function to check if a character is valid in a cookie name/value
// (simplified: alphanumeric only for this example)
pub open spec fn is_valid_cookie_char(c: u8) -> bool {
    (c >= 'a' as u8 && c <= 'z' as u8) ||
    (c >= 'A' as u8 && c <= 'Z' as u8) ||
    (c >= '0' as u8 && c <= '9' as u8)
}

// Check if all characters in a byte vector are valid
pub open spec fn all_valid_chars(s: Seq<u8>) -> bool {
    forall|i: int| 0 <= i < s.len() ==> #[trigger] is_valid_cookie_char(s[i])
}

// Serialize a single cookie to "key=value" format
fn serialize_cookie(key: &Vec<u8>, value: &Vec<u8>) -> (result: Vec<u8>)
    requires
        key.len() > 0,
        value.len() > 0,
        all_valid_chars(key@),
        all_valid_chars(value@),
    ensures
        result.len() == key.len() + 1 + value.len(),
{
    let mut result: Vec<u8> = Vec::new();
    
    // Copy key
    let mut i: usize = 0;
    while i < key.len()
        invariant
            i <= key.len(),
            result.len() == i,
    {
        result.push(key[i]);
        i = i + 1;
    }
    
    // Add '=' separator
    result.push('=' as u8);
    
    // Copy value
    let mut j: usize = 0;
    while j < value.len()
        invariant
            j <= value.len(),
            result.len() == key.len() + 1 + j,
    {
        result.push(value[j]);
        j = j + 1;
    }
    
    result
}

// Simple proof that serialized length equals key+1+value
proof fn lemma_serialize_length(key_len: nat, value_len: nat)
    ensures key_len + 1 + value_len > key_len,
{
    // Trivial arithmetic proof - the serialized form is longer than just the key
}

// Future extensions could include:
// - Parsing "key=value" strings back into Cookie structs
// - Handling multiple cookies separated by "; "
// - Validating cookie size limits (RFC6265 constraints)
// - Supporting cookie attributes like Path, Domain, Expires

} // verus!
\end{lstlisting}
\end{codebox}

\paragraph{Frama-C.} Sampled repository: \texttt{edwtjo/nixitch}.

\begin{promptbox}{Frama-C example --- sampled README}
\begin{lstlisting}[style=promptstyle]
Repository: edwtjo/nixitch

# Project moved!

The main repo is at [NixOS](https://github.com/NixOS/nix-idea)
\end{lstlisting}
\end{promptbox}

\begin{promptbox}{Frama-C example --- sampled reference snippet 1}
\begin{lstlisting}[style=promptstyle]
--- loop-invariant ---
A `loop invariant` is a property that holds before the loop and is preserved by every iteration. WP proves it on entry (establishment) and across one arbitrary iteration (preservation), then uses it together with the negated loop test to reason about the state after the loop. Strong enough invariants are the key to verifying loops.

```c
/*@
  requires n >= 0;
  assigns  \nothing;
  ensures  \result == n;
*/
int count_up(int n)
{
  int i = 0;
  /*@
    loop invariant 0 <= i <= n;
    loop assigns i;
    loop variant n - i;
  */
  while (i < n) {
    i++;
  }
  return i;
}
```
\end{lstlisting}
\end{promptbox}

\begin{promptbox}{Frama-C example --- sampled reference snippet 2}
\begin{lstlisting}[style=promptstyle]
--- assertion ---
An `//@ assert` annotation states a predicate that must hold at a specific program point inside a function body. WP generates a goal to prove it from the facts available there, and the asserted fact then becomes available to subsequent reasoning. Assertions are useful as intermediate lemmas that guide the prover toward the postcondition.

```c
/*@
  requires 0 <= x <= 100;
  assigns  \nothing;
  ensures  \result == 2 * x;
*/
int double_it(int x)
{
  int y = x + x;
  //@ assert y == 2 * x;
  return y;
}
```
\end{lstlisting}
\end{promptbox}

\begin{codebox}{Frama-C example --- Qwen2.5-Coder-32B response (verifies successfully)}
\begin{lstlisting}[style=codestyle]
/*
 * Nix Expression Path Validator
 * 
 * Inspired by the Nix package manager's path expressions, this program
 * validates and manipulates simple path strings representing package derivations.
 * A valid path contains only alphanumeric characters, hyphens, and slashes,
 * and doesn't start/end with invalid characters.
 */

#include <stddef.h>

/*@ predicate is_valid_char(char c) =
      ('a' <= c <= 'z') || ('A' <= c <= 'Z') ||
      ('0' <= c <= '9') || c == '-' || c == '/';
*/

/*@ predicate valid_path{L}(char *p, integer len) =
      len >= 0 &&
      (\forall integer i; 0 <= i < len ==> is_valid_char(p[i])) &&
      (len == 0 || p[0] != '/') &&
      (len == 0 || p[len-1] != '/');
*/

/*@
  requires \valid_read(path + (0 .. len-1));
  requires len >= 0;
  assigns  \nothing;
  ensures  \result == 1 <==> valid_path(path, len);
  ensures  \result == 0 || \result == 1;
*/
int validate_nix_path(const char *path, int len)
{
  // Empty path is valid
  if (len == 0) {
    return 1;
  }

  // Check first character
  char first = path[0];
  if (first == '/') {
    return 0;
  }

  // Check last character
  char last = path[len - 1];
  if (last == '/') {
    return 0;
  }

  // Check all characters
  /*@
    loop invariant 0 <= i <= len;
    loop invariant \forall integer j; 0 <= j < i ==> is_valid_char(path[j]);
    loop assigns i;
    loop variant len - i;
  */
  for (int i = 0; i < len; i++) {
    char c = path[i];
    int valid = ((c >= 'a' && c <= 'z') ||
                 (c >= 'A' && c <= 'Z') ||
                 (c >= '0' && c <= '9') ||
                 c == '-' || c == '/');
    
    if (!valid) {
      return 0;
    }
  }

  return 1;
}

/*@
  requires \valid_read(path + (0 .. len-1));
  requires len >= 0;
  requires start >= 0 && start <= len;
  assigns  \nothing;
  ensures  \result >= start && \result <= len;
  ensures  \forall integer i; start <= i < \result ==> path[i] != '/';
  ensures  \result < len ==> path[\result] == '/';
  ensures  \result == len ==> (\forall integer i; start <= i < len ==> path[i] != '/');
*/
int find_next_slash(const char *path, int len, int start)
{
  /*@
    loop invariant start <= i <= len;
    loop invariant \forall integer j; start <= j < i ==> path[j] != '/';
    loop assigns i;
    loop variant len - i;
  */
  for (int i = start; i < len; i++) {
    if (path[i] == '/') {
      return i;
    }
  }
  return len;
}

/* Future extensions:
 * - Count path components (separated by slashes)
 * - Validate specific Nix attribute names (alphanumeric, hyphen, underscore)
 * - Parse and validate full Nix expression paths
 */
\end{lstlisting}
\end{codebox}

\begin{figure}
    \centering
    \includegraphics[width=\linewidth]{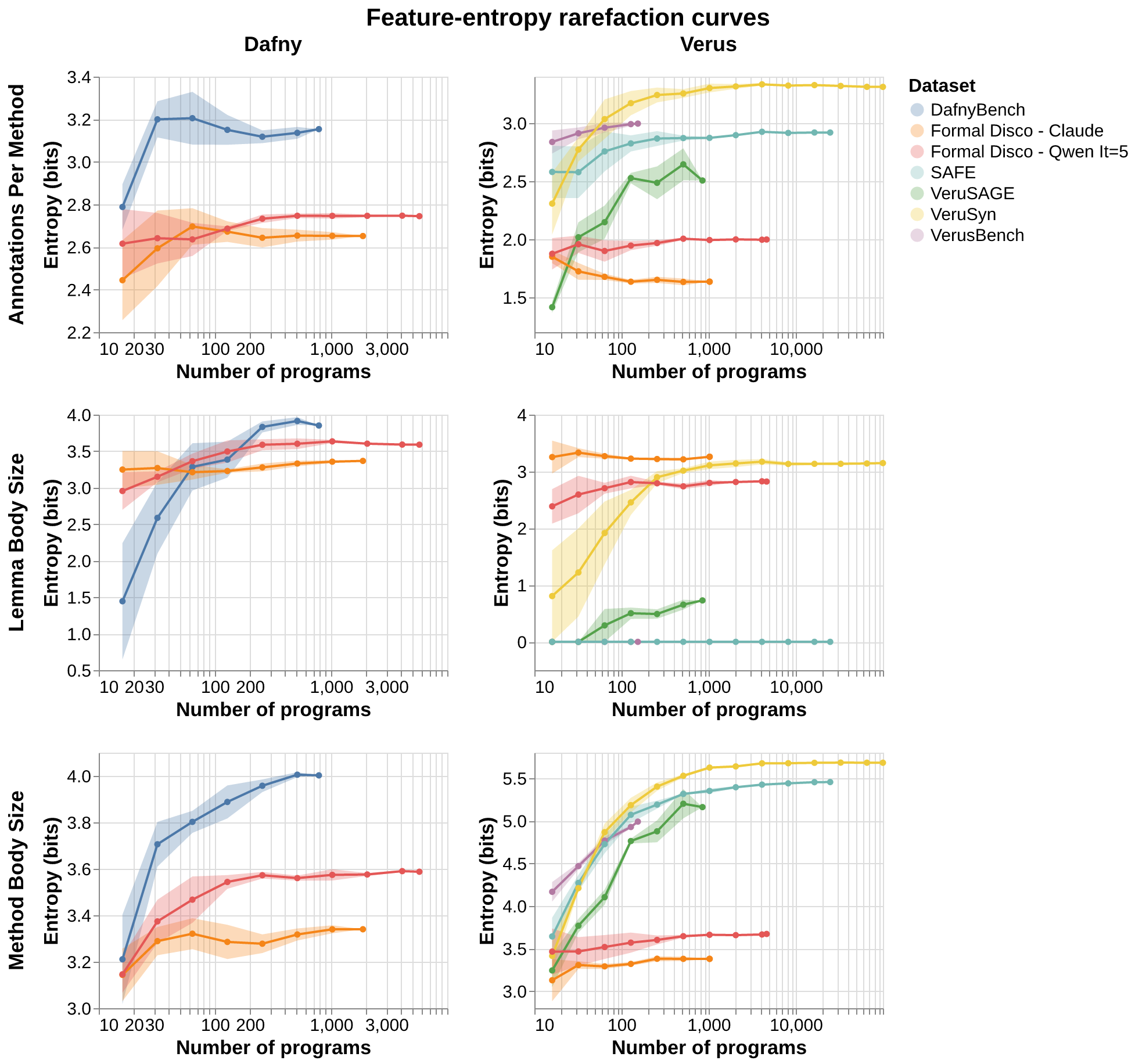}
    \caption{Rarefaction curves across (1) Number of logical annotations per method, (2) Proof/Body length of lemmas, and (3) Body length of methods (both in lines of code). All three are features included in entropy maximization}
    \label{fig:rarefaction-comparison-features2}
\end{figure}

\begin{figure}
    \centering
    \includegraphics[width=\linewidth]{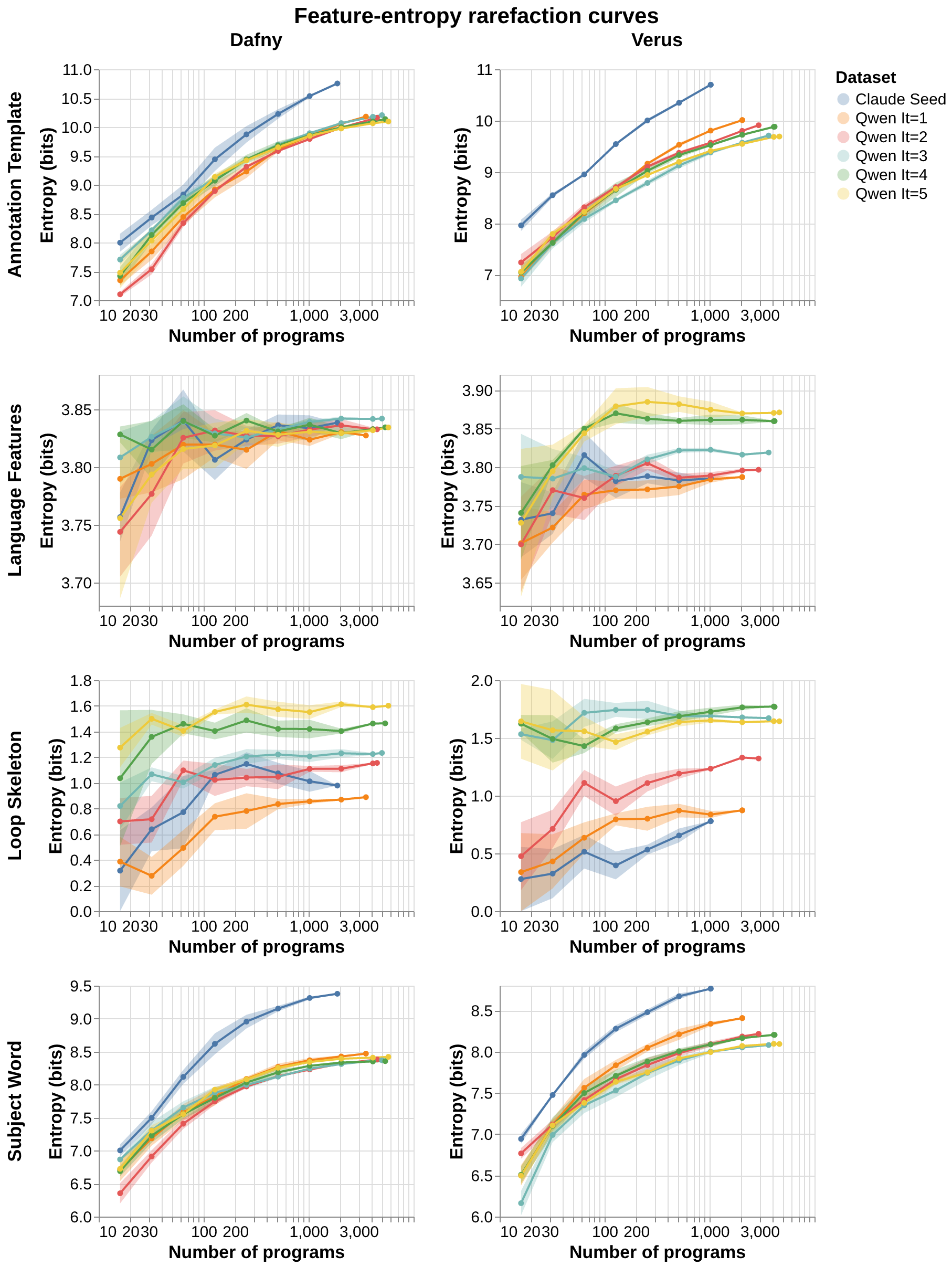}
    \caption{Rarefaction curves across self-improvement iterations, in the same features shown in Figure~\ref{fig:rarefaction-comparison-features1}. In features we do not explicitly optimize for, the initial \systemname{} runs with Claude still turn out more diverse; in cases where we do optimize (here, Loop Skeleton), our fine-tuned models take the lead.}
    \label{fig:rarefaction-self-improvement-1}
\end{figure}

\begin{figure}
    \centering
    \includegraphics[width=\linewidth]{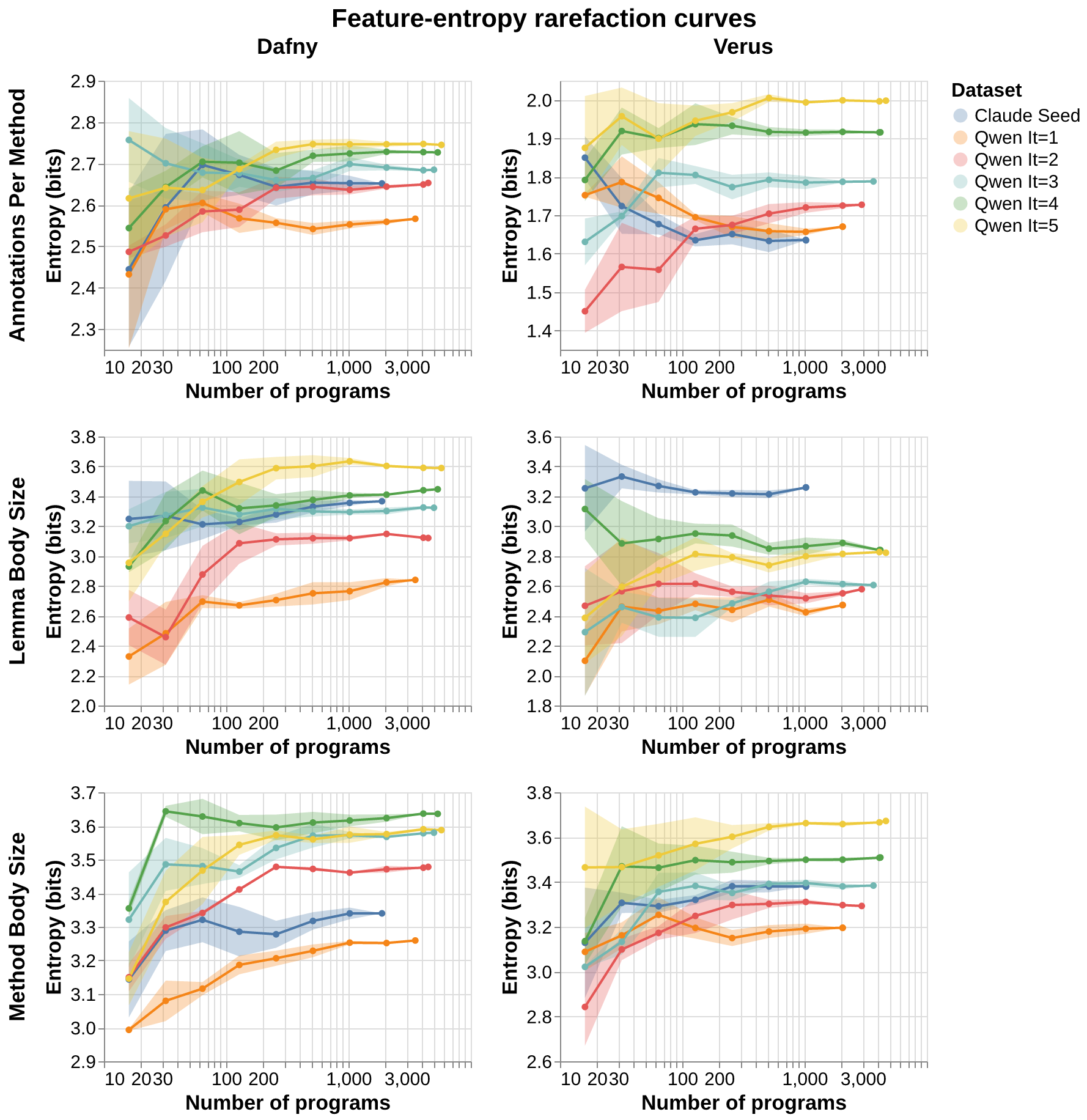}
    \caption{Rarefaction curves across self-improvement iterations, in the same features shown in Figure~\ref{fig:rarefaction-comparison-features2}.}
    \label{fig:rarefaction-self-improvement-2}
\end{figure}

\end{document}